\documentclass{article}
\makeatletter
\def\input@path{{arxiv/}{./}}
\makeatother
\usepackage{iclr2027_conference,times}
\usepackage{amsmath,amsfonts,bm}

\def\eqref#1{equation~\ref{#1}}
\def\1{\bm{1}}

\DeclareMathAlphabet{\mathsfit}{\encodingdefault}{\sfdefault}{m}{sl}
\SetMathAlphabet{\mathsfit}{bold}{\encodingdefault}{\sfdefault}{bx}{n}

\usepackage{graphicx}
\usepackage{booktabs}
\usepackage{array}
\usepackage{multirow}
\usepackage{longtable}
\usepackage{tabularx}
\newcolumntype{L}[1]{>{\raggedright\arraybackslash}p{#1}}
\newcolumntype{C}[1]{>{\centering\arraybackslash}p{#1}}
\newcolumntype{Y}{>{\centering\arraybackslash}X}
\usepackage{xcolor}
\usepackage{hyperref}
\usepackage{pax}
\makeatletter
\def\PAX@attrs@URI#1{#1}
\makeatother
\usepackage{wrapfig}
\usepackage{float}
\usepackage{url}
\usepackage{enumitem}
\newcommand{\mypara}[1]{\vspace{-1mm}\noindent\textbf{#1}}
\usepackage{tcolorbox}
\tcbuselibrary{listings,breakable}
\newtcblisting{promptbox}[2][]{%
  colback=gray!5!white,
  colframe=gray!80!black,
  fonttitle=\bfseries,
  title={#2},
  breakable,
  width=\linewidth,
  before=\noindent,
  left skip=0pt,
  right skip=0pt,
  boxrule=0.5pt,
  left=4pt,
  right=4pt,
  top=4pt,
  bottom=4pt,
  listing only,
  listing engine=listings,
  listing options={
    basicstyle=\ttfamily\scriptsize,
    breaklines=true,
    breakatwhitespace=false,
    columns=fullflexible,
    keepspaces=true,
    showstringspaces=false,
    literate={–}{{-}}1 {—}{{--}}2 {“}{{``}}2 {”}{{''}}2
  },
  #1
}
\usepackage{xspace}
\newcommand{\ours}{WebViSE\xspace}

\title{WebViSE: Scaling Visual Design Intelligence from the Open Web for LLM Website Generation}
\author{%
  Yifei Sun\textsuperscript{1,2,*}, Peng Zhao\textsuperscript{2,*},
  Le Zhang\textsuperscript{2}, Bingjin Chen\textsuperscript{2},
  Zhengfan Wu\textsuperscript{2}\\
  \normalfont\textsuperscript{1}Peking University\qquad
  \textsuperscript{2}Baidu Inc., Beijing, China\\
  \normalfont\textsuperscript{*}Equal contribution.
}

\date{}
\hypersetup{
  pdftitle={WebViSE: Scaling Visual Design Intelligence from the Open Web for LLM Website Generation},
  pdfkeywords={website generation, visual design, agent skills, skill retrieval, distillation}
}

\iclrfinalcopy
\begin{document}
\maketitle
\lhead{Preprint}
\vspace{-2\baselineskip}
\begin{abstract}
Large language models have made remarkable strides in website development, enabling complete and interactive web experiences to be built from natural-language instructions with far less effort. Yet, non-expert users often struggle to articulate sophisticated visual and interaction designs in natural language. Meanwhile, the open web already contains a vast reservoir of such design knowledge, embodied in mature website implementations. To this end, we introduce \textbf{Web}site-derived \textbf{Vi}sual \textbf{S}kill \textbf{E}xtraction (\textbf{WebViSE}), a framework that automatically abstracts these implementations into reusable Visual Skills and builds a structured Skill library. To scale the library, WebViSE organizes its Skills in a multi-parent DAG and jointly evolves Skill descriptions and DAG structure using Skill-conditioned Potential Queries, enabling efficient matching between diverse user instructions and relevant Skills. The auto-evolved structure outperforms non-hierarchical LLM retrieval baselines at only 39.09\% of the estimated retrieval cost with Qwen3.8-Max. For any models, WebViSE can be directly injected as plug-and-play guidance, enabling diverse models to produce more visually compelling and aesthetically refined web pages. For open-source models, WebViSE can be used to build a data-synthesis pipeline that internalizes visual design intelligence into model weights. Experiments across multiple web development benchmarks and human evaluation consistently show that WebViSE improves the visual quality and interaction fidelity of artifacts. Overall, our work establishes an automated path for extracting, organizing, and transferring design intelligence from the open web to website-building models. We release all extracted Visual Skills, the associated Skill DAG, and our code
at \url{https://github.com/YifeiSun-0710/WebViSE}.

\end{abstract}

\section{Introduction}
\label{sec:introduction}
\vspace{-2mm}
Large language models (LLMs) can increasingly generate complete and interactive
websites from natural-language instructions
\citep{lu2025webgenbench,yang2026cookiebench,dai2026iwebgenbench}. Yet these
instructions typically specify a website's purpose, content, and functionality
while leaving its visual presentation and interaction design underspecified.
Non-expert users often
lack the design knowledge and vocabulary to fill this gap, leaving Website
Generators to fall back on familiar patterns and produce conventional,
homogeneous designs.
\vspace{-2mm}

The open web offers a rich source of visual design intelligence, embodied in
mature websites through diverse visual and interaction mechanisms. Yet this
knowledge is entangled with source-specific content, branding, assets, layouts,
and implementation details, making it difficult to transfer directly to a new
request. Such transfer requires identifying the underlying mechanism,
abstracting it from its original context, and expressing it in a form that a
Website Generator can apply. Recent work represents reusable procedural
knowledge as Agent Skills that models can load and apply across tasks
\citep{anthropic2025agentskills,pan2026anything2skill,mi2026skillpro,
liang2026skillstructure}. We extend this paradigm to visual design with
Website-Derived Visual Skills, which recover concrete visual and interaction
mechanisms from implemented websites with the conditions under which
they are applicable.

\begin{figure}[t]
    \centering
    \includegraphics[width=0.88\linewidth]{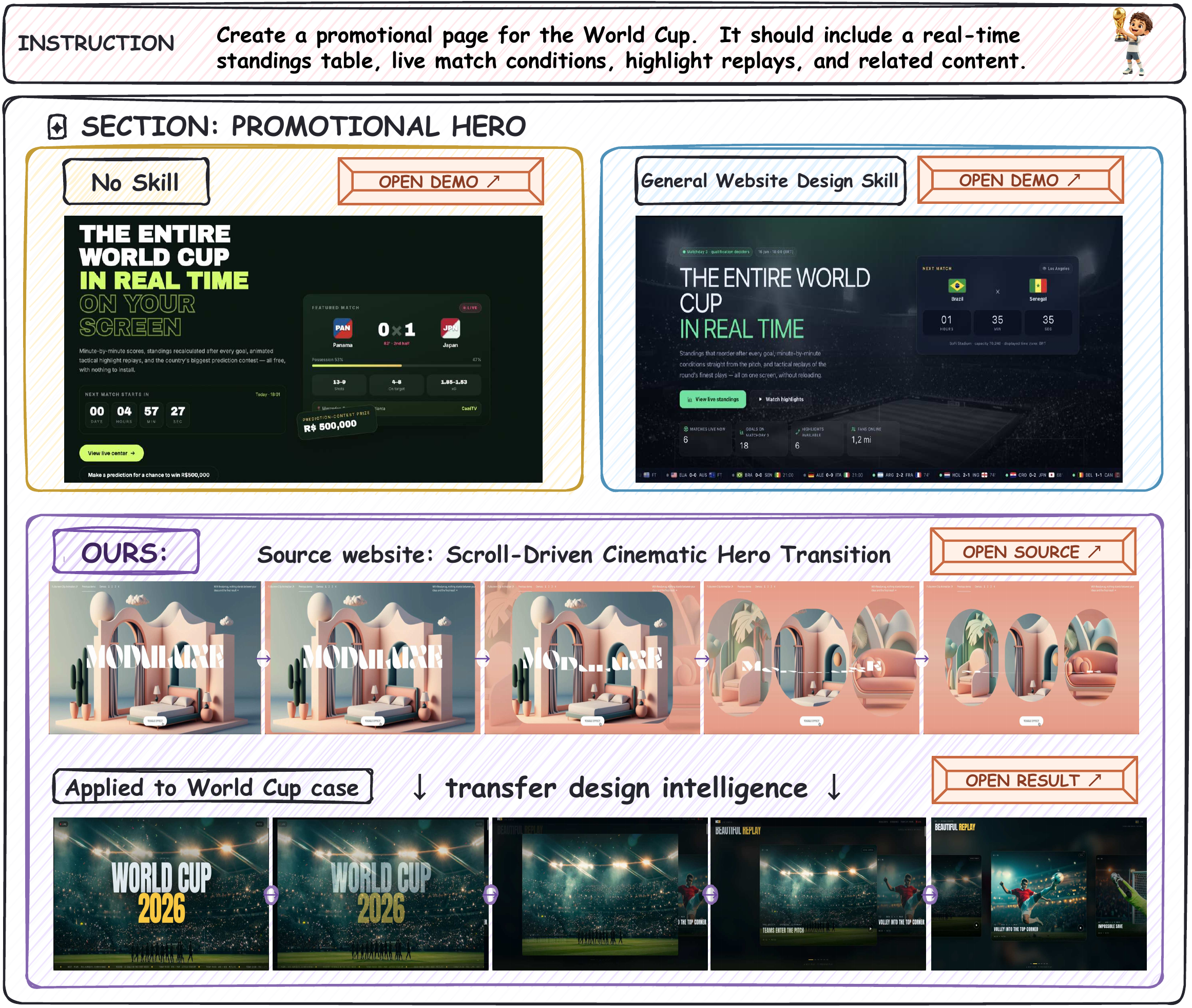}
    \vspace{-1\baselineskip}
    \caption{\textbf{Our Website-Derived Visual Skills enrich website experiences.}
No Skill and General Website Design Skill produce similar World Cup pages, whereas
ours enhance their visual and interactive design with
mechanisms recovered from source websites; \textbf{For the full interactive experience, we encourage readers to  CLICK the $\nearrow$ icons to open webpages.}}
    \label{fig:skill_usage}
    \vspace{-1.5\baselineskip}
    
\end{figure}

\vspace{-1mm}

We first test whether mechanisms recovered from implemented websites can
transfer to a strong Website Generator. As shown in
Figure~\ref{fig:skill_usage}, using the same instruction and Claude Opus
5~\citep{anthropic2026opus5}, we compare No Skill, General Website Design
Skill (a popular open-source design-intelligence
repository UI/UX Pro Max Skill~\citep{nextlevelbuilder2026uiuxpromax}), and Website-Derived Visual Skills. No Skill and the General
Website Design Skill produce similar designs, indicating that generic design
guidance is largely redundant for a strong generator. Website-Derived Visual
Skills instead introduce a scroll-driven hero transition absent from the original instruction, motivating the transfer of
mechanism-level design intelligence from implemented websites.

\vspace{-1mm}

Turning this observation into a practical system requires scaling from individual
mechanisms to a Skill library built from vast collections of well-designed
websites. Flat LLM retrieval inspects every Skill
for each request, so token cost grows with library size. A hierarchy with
one parent per Skill cannot capture its multiple design roles, and manually
curating Skill descriptions and routes does not scale. A scalable
solution must therefore support selective access, preserve multiple retrieval
paths, and automatically refine both the Skills and their organization.

\vspace{-1mm}

To address these challenges, we introduce \textbf{Web}site-derived
\textbf{Vi}sual \textbf{S}kill \textbf{E}xtraction (\textbf{WebViSE}), a
framework for extracting, organizing, and transferring visual design
intelligence from the open web. WebViSE abstracts visual and interaction
mechanisms from website implementations into reusable Visual Skills and
organizes them in a multi-parent Skill DAG. Multi-path routing and applicability
review provide selective access to relevant Skills, while Skill-conditioned
Potential Queries provide automatic supervision for jointly evolving the Skill
descriptions and DAG structure. The evolved library can directly augment any
Website Generator, including closed-source models, as plug-and-play guidance.
For open-source models, WebViSE can further synthesize data for supervised
fine-tuning (SFT), internalizing visual design intelligence into model weights
and enabling generation from the original instruction without inference-time
Skill retrieval. Our contributions are:
\begin{itemize}[leftmargin=3mm]
\vspace{-2.5mm}
\vspace{-0.5mm}\item We introduce an automated approach for extracting mechanism-level
Visual Skills from implemented websites while abstracting away
source-specific content and presentation.
\vspace{-2mm}
\item We propose a multi-parent Skill DAG with query-induced joint evolution of
Skill documents and routing structure, supporting scalable and cost-efficient retrieval from a growing Skill library.
\vspace{-2mm}
\vspace{-0.5mm}\item We enable visual design intelligence to be transferred through both
plug-and-play guidance and Skill-augmented training, and validate the
resulting improvements across diverse Website Generators, benchmarks, and
human evaluation.
\vspace{-2mm}
\end{itemize}

\section{Method}
\begin{figure*}[t]
    \centering
    \includegraphics[width=\textwidth]
    {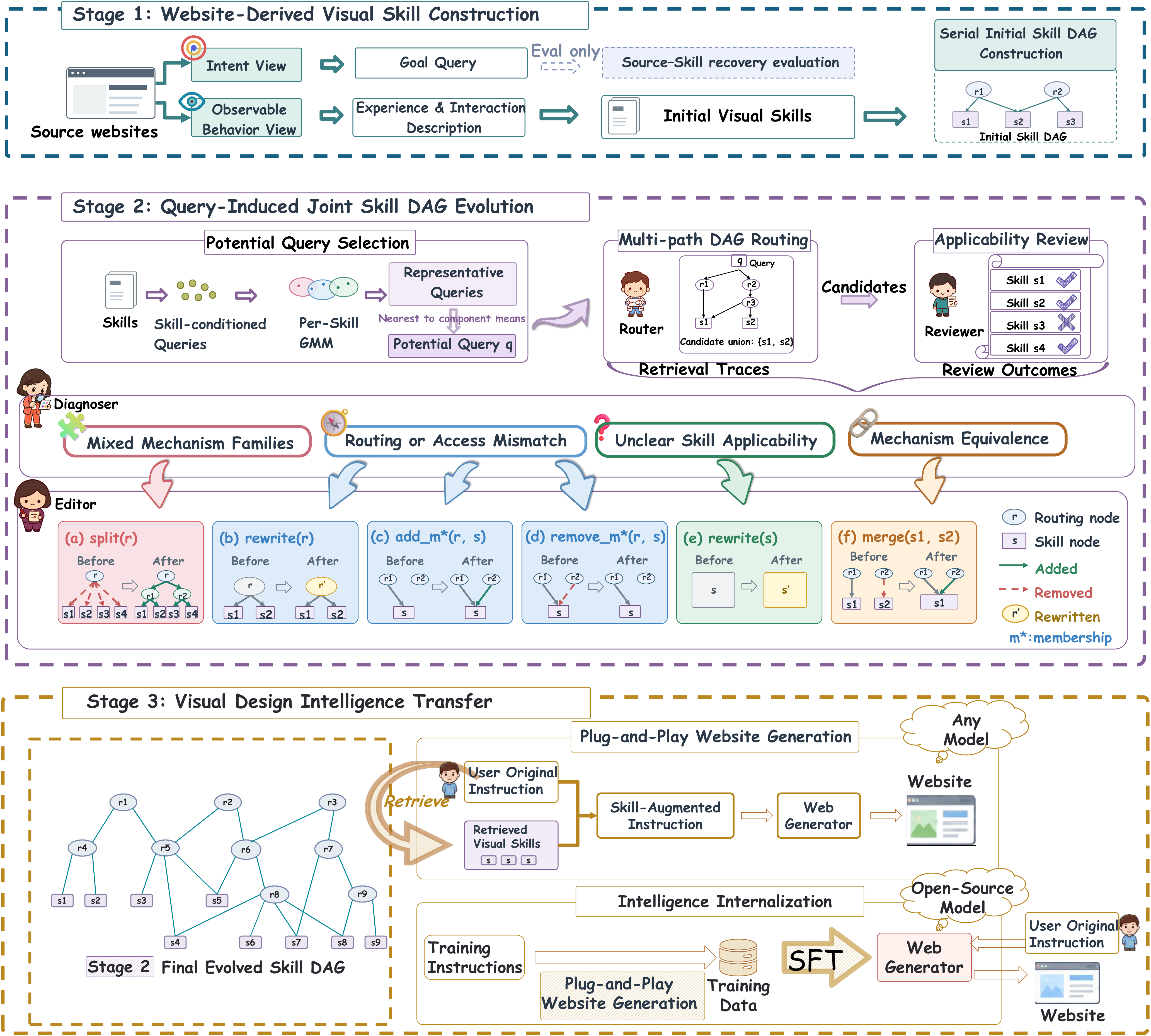}
    \vspace{-2\baselineskip}
    \caption{\textbf{Overview of WebViSE.} WebViSE constructs Visual Skills from source websites, jointly evolves the Skill descriptions and a multi-parent routing DAG using Skill-conditioned Potential Queries, and then transfers the resulting visual design intelligence.}
     \vspace{-1.5\baselineskip}
    \label{fig:method_overview}
\end{figure*}

\label{sec:method}
WebViSE transforms design knowledge embedded in source websites into reusable
Visual Skills and transfers this visual design intelligence to Website
Generators. As illustrated in Figure~\ref{fig:method_overview}, Stage 1
constructs Visual Skills from complementary views of source websites and
integrates them into an initial Skill DAG. Stage 2 jointly evolves the Skill
descriptions and multi-parent DAG using Skill-conditioned Potential Queries to
support scalable Skill access, replacing full-library inspection with selective
routing to relevant Skills. Stage 3 transfers the evolved Skills through
plug-and-play augmentation or intelligence internalization.
 
\subsection{Website-Derived Visual Skill Construction}
\label{subsec:skill_construction}
 
\mypara{Source Views and Initial Skill Synthesis.}
Inspired by the ISO~9241 family, which covers user goals, user experience,
and human--system interaction
\citep{iso9241_11_2018,iso9241_210_2019,iso9241_110_2020},
we derive two complementary views from each source website's implementation:
\begin{itemize}[leftmargin=2mm]
\vspace{-2mm}
    \item \textbf{Intent View}
    $\rightarrow$ \textbf{Goal Query}.
    The Goal Query captures the website's purpose and foreground content without visual
    or interaction details. It informs candidate Skills' usage contexts. We also use it for source-Skill recovery evaluation.

\vspace{-2mm}
    \item \textbf{Observable Behavior View}
    $\rightarrow$ \textbf{Experience \& Interaction Descriptions}.
    The Experience Description captures the dominant user journey and
    perceptual outcome. The Interaction Description records relevant inputs,
    affected objects, state changes, and visible feedback. Together, they
    supply the observable experience from which candidate Skills are distilled.
\vspace{-2mm}
\end{itemize}
We distill candidate Visual Skills from both views: Goal informs where a Skill
applies, while Experience and Interaction characterize its observable
behavior. Given a selected candidate and the website's source code, the final
constructor recovers transferable implementation mechanisms and writes an
initial Visual Skill $s_i^{(0)}$. 

\begin{wrapfigure}{r}{0.65\textwidth}
    \centering
    \vspace{-1.5\baselineskip}
    \includegraphics[width=1.0\linewidth]
{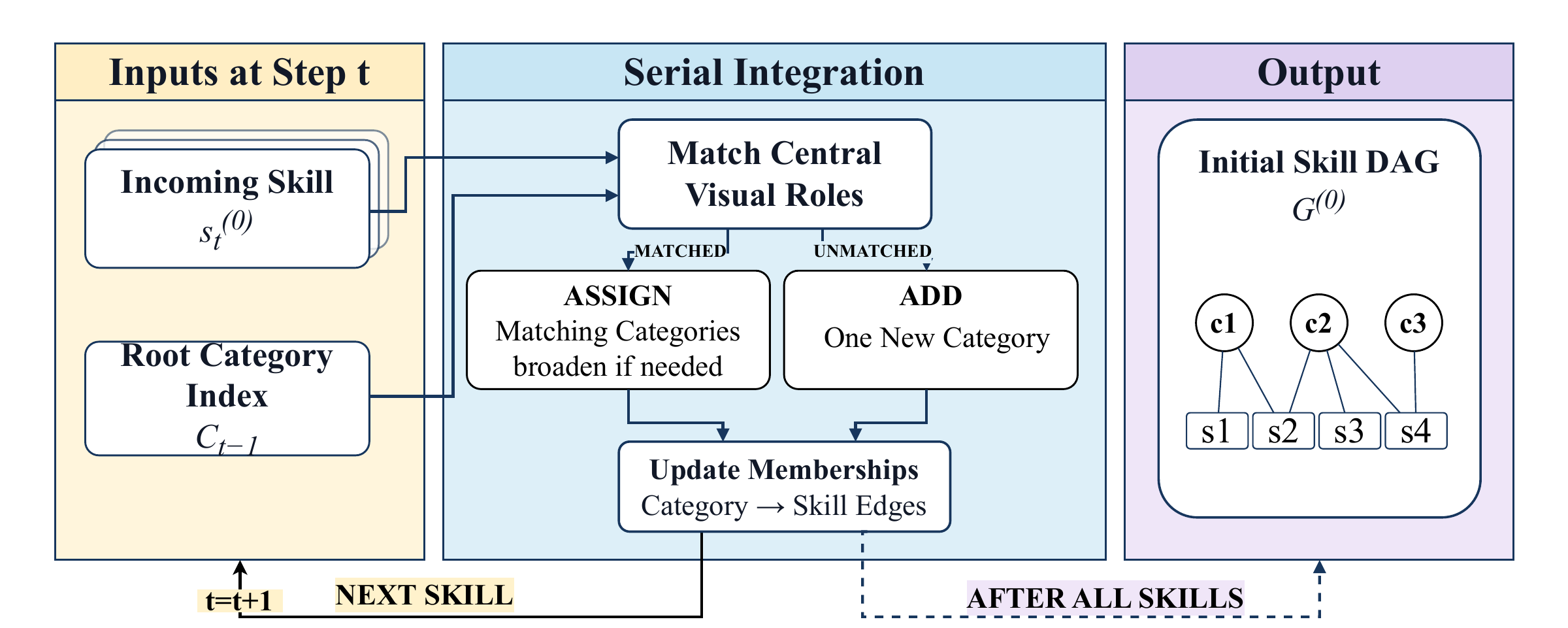}
    \vspace{-1.5\baselineskip}
    \caption{\textbf{Serial construction of the initial one-level Skill DAG.}}
    \label{fig:initial_skill_dag_construction}
    \vspace{-1.5\baselineskip}
\end{wrapfigure}

\mypara{Serial Initial Skill DAG Construction.}
Figure~\ref{fig:initial_skill_dag_construction} illustrates how we construct the
initial Skill DAG by integrating one initial Skill at a time. Let
$\mathcal{S}^{(0)}=\{s_1^{(0)},\ldots,s_N^{(0)}\}$ denote the set of initial
Visual Skills. At step $t$, the model receives the incoming Skill
$s_t^{(0)}$ and the current root Category index $\mathcal{C}_{t-1}$. It first
identifies the central visual roles of the incoming Skill. For each matched
role, the Skill is assigned to all existing Categories whose descriptions cover
that role. If a Category is too narrow, its description may be broadened
without changing its identity. If the most salient role remains unmatched, the
model adds at most one new Category for that role. It then updates the
Category-to-Skill memberships and proceeds to the next Skill.
After all initial Skills have been integrated, the resulting graph is
\begin{equation}
G^{(0)}=(V^{(0)},E^{(0)}), \qquad
V^{(0)}=\mathcal{C}^{(0)}\mathbin{\dot\cup}\mathcal{S}^{(0)}, \qquad
E^{(0)}\subseteq\mathcal{C}^{(0)}\times\mathcal{S}^{(0)}.
\end{equation}
Here, $\mathcal{C}^{(0)}$ denotes the root Categories,
$\mathcal{S}^{(0)}$ denotes the initial Visual Skills, and
$E^{(0)}$ contains the directed Category-to-Skill membership edges. The
membership relation is many-to-many: a Category may contain multiple Skills,
and a Skill may belong to multiple Categories. Thus, $G^{(0)}$ is a one-level bipartite DAG.
In the initial graph, Categories are root routing nodes. During later Skill DAG
evolution, split operations may introduce routing nodes below these roots. We use \emph{routing node} as the general term for any non-Skill node used to
route Queries. The resulting $(\mathcal{S}^{(0)},G^{(0)})$ provides the initial
state for query-induced Skill DAG evolution.

\subsection{Query-Induced Joint Skill DAG Evolution}
\label{subsec:joint_optimization}

As the Skill library grows, inspecting every Skill for each Query becomes
increasingly costly. WebViSE instead uses Skill-conditioned Potential Queries
to jointly evolve Skill descriptions and a multi-parent DAG, supporting
scalable Skill access through selective multi-path routing.

\subsubsection{Skill-Conditioned Potential Query Selection}

For each initial Skill $s_i^{(0)}$, we synthesize Potential Queries$\mathcal{Q}_i$ from its name and description. Each Query represents a plausible website request that the Skill is intended to support. Because the source Skill is known by construction, each pair $(q,s_i^{(0)})$ provides automatic supervision for source-Skill recovery. To reduce redundancy while preserving coverage of distinct use cases, we embed the Queries with an encoder $\phi$ and fit a per-Skill Gaussian mixture model:
\begin{equation}
p_i(\mathbf{z})
=
\sum_{m=1}^{K_i}
\pi_{im}\,
\mathcal{N}
\left(
\mathbf{z};
\boldsymbol{\mu}_{im},
\boldsymbol{\Sigma}_{im}
\right),
\qquad
\mathbf{z}=\phi(q).
\end{equation}
Each mixture component captures a semantic mode of the Skill's Query space. We
select the generated Query closest to each component mean, yielding a compact
set of representative Query--Skill pairs. Each selected Query probes a distinct
use case, and failure to recover and retain its source Skill after DAG routing
and applicability review provides diagnostic evidence for refining the Skill
documents and DAG structure.
 \vspace{-1mm}

\subsubsection{Multi-Path DAG Routing and Applicability Review}
\label{subsec:skill_dag_retrieval}

For each representative Potential Query $q$, the \textbf{Router} recursively follows all
matching branches from the root routing nodes to Skill nodes. This multi-path
traversal preserves alternative semantic routes while restricting retrieval to
a query-specific subgraph. The reached Skills are deduplicated to form the
candidate set $\mathcal{S}_{\mathrm{cand}}(q)$, and the traversed nodes and
paths are recorded as retrieval traces.

The \textbf{Reviewer} evaluates candidates in $\mathcal{S}_{\mathrm{cand}}(q)$ using the
Query and their complete Skill documents. It removes candidates whose
mechanisms do not support the requested visual or interaction outcome, ranks
the remainder by applicability, and retains the top-$K$ results.

Restricting full-document inspection to routed candidates avoids repeated
full-library review, enabling scalable Skill access as the library grows. The
resulting retrieval traces and review outcomes are passed to the
\textbf{Diagnoser}, whose diagnoses guide the \textbf{Editor}.

 \vspace{-1mm}
\subsubsection{Failure Diagnosis and Joint Skill DAG Editing}
\label{subsec:joint_editing}

For each Potential Query, the \textbf{Diagnoser} compares the retrieval trace
and review outcome against the known reference Skill. It aggregates recurring
evidence across Queries into four failure types. The \textbf{Editor} maps these
diagnoses to corresponding atomic operations:
\begin{itemize}[leftmargin=4mm]
\vspace{-1\baselineskip}
\item Mixed mechanism families: a routing node contains Skills with
incompatible visual or interaction roles. The Editor applies
\texttt{split} to separate these families and redistribute their
memberships.
\vspace{-0.5\baselineskip}\item  Routing or access mismatch: the reference Skill is unreachable
in the current DAG, or an inaccurate routing description directs the Query
to an incorrect branch. The Editor applies \textit{rewrite}, \textit{add\_membership}, or \textit{remove\_membership} to repair the affected routing node or connection.
\vspace{-0.5\baselineskip}\item Unclear Skill applicability: a Skill document does not explain how its mechanism supports the Query. The Editor applies \textit{rewrite} to clarify its applicability while keeping its central mechanism.
\vspace{-1.5\baselineskip}\item Mechanism equivalence: Skills describe equivalent reusable mechanisms. The Editor proposes \textit{merge}, which is accepted only after equivalence
is confirmed using the complete Skill documents.
\vspace{-1.5\baselineskip}
\end{itemize}

The \textbf{Editor} applies the accepted operations to jointly update the Skill
documents and DAG. Subsequent Potential Queries exercise the revised library,
closing the evolution loop and exposing remaining retrieval failures. After
evolution concludes, the resulting Skill library and DAG support both
plug-and-play retrieval and training-data construction for intelligence
internalization.

 \vspace{-1mm}
\subsection{Visual Design Intelligence Transfer}
\label{subsec:visual_design_intell}
 
\mypara{Plug-and-Play Website Generation.} WebViSE provides plug-and-play augmentation for any Web Generator, including
closed-source models. Given an original user instruction, WebViSE retrieves
applicable Visual Skills from the evolved DAG through multi-path routing and
applicability review. The retrieved Skills are combined with the original
instruction for generation.

\mypara{Intelligence Internalization.}
For open-source models, WebViSE further internalizes visual design intelligence through SFT. The same plug-and-play pipeline augments each training instruction with retrieved Visual Skills, from which a teacher Web Generator synthesizes the website implementation. The implementation is then paired with the original instruction to train the target model. At inference time, the target model receives only the original instruction, yet can apply the internalized visual design intelligence without Skill retrieval.

\vspace{-2mm} 
\section{Experimental Evaluation of WebViSE for Website Generation}
\label{sec:web_generation_evaluation}
\vspace{-2mm}
We evaluate WebViSE across diverse web-generation benchmarks using a broad set of LLMs, ranging from a 27B open-weight model to closed-source APIs. Together, the plug-and-play evaluation, controlled ablations,
Skill-augmented training, and blinded human evaluation answer four questions:
\textbf{(1)} Does \ours provide plug-and-play improvements across diverse
Website Generators and benchmarks? \textbf{(2)} What makes \ours effective as
plug-and-play guidance? \textbf{(3)} Are the improvements produced by \ours
perceptible to human evaluators? \textbf{(4)} Can a smaller model internalize
visual design intelligence from \ours and apply it without inference-time Skill
retrieval?
\vspace{-1mm}
\vspace{-2mm}
\subsection{Experimental Setup}
\label{sec:web_generation_setup}

\mypara{Visual Skill library and retrieval pipeline.}
We extract 864 Visual Skills from 850 publicly accessible source websites and
organize them into a multi-parent Skill DAG. In all website-generation
experiments, Claude Opus 5 retrieves Visual Skills using the same frozen,
evolved Skill DAG and retrieval pipeline. The retrieved Skills are supplied
to each Web Generator, which produces its own website. These experiments
test the benefit of supplying retrieved Visual Skills. Separately, the
retrieval experiments evaluate whether the DAG reduces retrieval cost while
maintaining competitive retrieval quality, as described in
Section~\ref{sec:skill_retrieval_evaluation}. The same retrieval pipeline is
used to construct the Skill-augmented training data in
Section~\ref{subsec:capability_internalization}.
Appendix~\ref{app:setup_details} provides details on the data sources,
filtering procedure, category and graph statistics, and Potential-Query
construction.

\mypara{Evaluation Protocol and Benchmarks.}
For each task, the \textsc{Base} condition generates a website from the original task instruction, whereas \textsc{WebViSE} augments the same instruction with the retrieved Visual Skills. Whenever an evaluation component requires the task intent, it receives the original task instruction only. This keeps the evaluation grounded in the original task intent and requirements. We evaluate website generation on three complementary benchmarks. Detailed benchmark configurations are provided in Appendix~\ref{app:benchmarks}.
\begin{itemize}[leftmargin=3mm]
\vspace{-2mm}
\vspace{-0.5mm}\item \underline{WebGen-Bench}~\citep{lu2025webgenbench}
measures the functional correctness of multi-file websites using UI test cases
that pair browser operations with expected outcomes. We use all 101 instructions
(647 test cases) and report Accuracy, assigning full credit to \textsc{yes} and
half credit to \textsc{partial}.

\vspace{-2mm}
\item \underline{MiniAppBench}~\citep{zhang2026miniappbench}
focuses on open-ended HTML applications requiring both visual design and custom interaction logic. We use its 100-task validation split. For readability and consistent presentation across benchmarks, we rescale the original $[0,1]$ scores by 100 and report Intention, Static, and Dynamic on a 0--100 scale.

\vspace{-2mm}
\item \underline{Cookie-Bench}~\citep{yang2026cookiebench}
covers both static pages and interactive applications. We evaluate its 844-task
website subset in dynamic mode, where the evaluator interacts with each rendered
website before assigning Functionality, Aesthetics, and Total scores.
\vspace{-2mm}
\end{itemize}

 \vspace{-1mm}
\subsection{Plug-and-Play Website Generation}
\label{sec:plugin_results}

\subsubsection{Main Results across Models and Benchmarks}
\vspace{-1mm}
\begin{table}[h]
    \centering
    \caption{\textbf{Plug-and-play website generation results across models and
    benchmarks.} Bold denotes the best score per column; underline denotes the
    second best. }
    \label{tab:web_generation_plugin_results}
    \footnotesize
    \setlength{\tabcolsep}{0.8pt}
    \renewcommand{\arraystretch}{1.0}
    \newcommand{\wvgain}[1]{\raisebox{0.15ex}{\scalebox{0.60}{\textcolor{green!45!black}{$\uparrow$#1}}}}
    \newcommand{\wvloss}[1]{\raisebox{0.15ex}{\scalebox{0.60}{\textcolor{red!75!black}{$\downarrow$#1}}}}
    \newcommand{\basevalue}[1]{\makebox[3.45em][l]{\makebox[2.20em][r]{#1}}}
    \newcommand{\wvscore}[2]{\makebox[3.45em][l]{\makebox[2.20em][r]{#1}\makebox[1.25em][l]{#2}}}
    \resizebox{0.95\linewidth}{!}{%
    \begin{tabularx}{\linewidth}{@{}L{0.21\linewidth}L{0.115\linewidth}@{\hspace{2pt}}C{0.13\linewidth}*{6}{Y}@{}}
        \toprule
        \multirow{2}{=}{Model} & \multirow{2}{*}{Setting}
        & \multicolumn{1}{c}{WebGen-Bench}
        & \multicolumn{3}{c}{MiniAppBench}
        & \multicolumn{3}{c}{Cookie-Bench} \\
        \cmidrule(lr){3-3}\cmidrule(lr){4-6}\cmidrule(lr){7-9}
        & & \multicolumn{1}{c}{\mbox{Accuracy (\%)}} & Int.  & Stat.  & Dyn.
        & Func. & Aes. & Total \\
        \midrule
        \multirow{2}{=}{Claude Opus 5\\~\citep{anthropic2026opus5}}
        & Base & \basevalue{56.80} & \basevalue{\underline{95.91}} & \basevalue{\underline{85.94}} & \basevalue{\underline{78.23}} & \basevalue{\textbf{97.81}} & \basevalue{95.46} & \basevalue{\underline{94.87}} \\
        & \textsc{+WebViSE} & \wvscore{\textbf{60.05}}{\wvgain{3.25}} & \wvscore{95.70}{\wvloss{0.21}} & \wvscore{\textbf{86.44}}{\wvgain{0.50}} & \wvscore{\textbf{80.40}}{\wvgain{2.17}} & \wvscore{\underline{96.74}}{\wvloss{1.07}} & \wvscore{\textbf{96.52}}{\wvgain{1.06}} & \wvscore{\textbf{94.99}}{\wvgain{0.12}} \\
        \midrule
        \multirow{2}{=}{Kimi K3\\~\citep{kimiteam2026kimik3}}
        & Base & \basevalue{55.26} & \basevalue{95.78} & \basevalue{82.05} & \basevalue{73.30} & \basevalue{94.68} & \basevalue{94.55} & \basevalue{91.89} \\
        & \textsc{+WebViSE} & \wvscore{55.41}{\wvgain{0.15}} & \wvscore{95.80}{\wvgain{0.02}} & \wvscore{82.60}{\wvgain{0.55}} & \wvscore{73.42}{\wvgain{0.12}} & \wvscore{96.09}{\wvgain{1.41}} & \wvscore{92.95}{\wvloss{1.60}} & \wvscore{92.78}{\wvgain{0.89}} \\
        \midrule
        \multirow{2}{=}{Qwen3.8-Max\\~\citep{qwen2026qwen38max}}
        & Base & \basevalue{53.48} & \basevalue{95.34} & \basevalue{81.97} & \basevalue{74.16} & \basevalue{95.60} & \basevalue{96.24} & \basevalue{93.51} \\
        & \textsc{+WebViSE} & \wvscore{55.72}{\wvgain{2.24}} & \wvscore{95.64}{\wvgain{0.30}} & \wvscore{81.50}{\wvloss{0.47}} & \wvscore{74.95}{\wvgain{0.79}} & \wvscore{96.00}{\wvgain{0.40}} & \wvscore{\underline{96.27}}{\wvgain{0.03}} & \wvscore{93.71}{\wvgain{0.20}} \\
        \midrule
        \multirow{2}{=}{GLM-5.3\\~\citep{zhipuai2026glm53}}
        & Base & \basevalue{50.39} & \basevalue{93.32} & \basevalue{80.57} & \basevalue{70.92} & \basevalue{94.17} & \basevalue{92.82} & \basevalue{90.01} \\
        & \textsc{+WebViSE} & \wvscore{47.60}{\wvloss{2.79}} & \wvscore{95.58}{\wvgain{2.26}} & \wvscore{82.92}{\wvgain{2.35}} & \wvscore{71.85}{\wvgain{0.93}} & \wvscore{94.19}{\wvgain{0.02}} & \wvscore{91.55}{\wvloss{1.27}} & \wvscore{89.73}{\wvloss{0.28}} \\
        \midrule
        \multirow{2}{=}{DeepSeek-V4.1-Flash\\~\citep{deepseekai2026deepseekv41flash}}
        & Base & \basevalue{58.11} & \basevalue{95.86} & \basevalue{85.13} & \basevalue{74.64} & \basevalue{95.30} & \basevalue{91.63} & \basevalue{90.38} \\
        & \textsc{+WebViSE} & \wvscore{\underline{59.27}}{\wvgain{1.16}} & \wvscore{\textbf{96.36}}{\wvgain{0.50}} & \wvscore{84.48}{\wvloss{0.65}} & \wvscore{76.76}{\wvgain{2.12}} & \wvscore{95.60}{\wvgain{0.30}} & \wvscore{92.67}{\wvgain{1.04}} & \wvscore{91.45}{\wvgain{1.07}} \\
        \midrule
        \multirow{2}{=}{GPT-5.6-Sol Standard\\~\citep{openai2026gpt56}}
        & Base & \basevalue{46.29} & \basevalue{93.84} & \basevalue{80.48} & \basevalue{73.71} & \basevalue{91.92} & \basevalue{91.13} & \basevalue{87.06} \\
        & \textsc{+WebViSE} & \wvscore{51.24}{\wvgain{4.95}} & \wvscore{94.83}{\wvgain{0.99}} & \wvscore{80.96}{\wvgain{0.48}} & \wvscore{74.16}{\wvgain{0.45}} & \wvscore{93.36}{\wvgain{1.44}} & \wvscore{92.32}{\wvgain{1.19}} & \wvscore{89.03}{\wvgain{1.97}} \\
        \midrule
        \multirow{2}{=}{Qwen3.8-27B \\~\citep{qwen2026qwen3827b}}
        & Base & \basevalue{52.86} & \basevalue{93.45} & \basevalue{79.06} & \basevalue{68.85} & \basevalue{93.98} & \basevalue{92.95} & \basevalue{90.13} \\
        & \textsc{+WebViSE} & \wvscore{54.40}{\wvgain{1.54}} & \wvscore{94.93}{\wvgain{1.48}} & \wvscore{80.63}{\wvgain{1.57}} & \wvscore{70.04}{\wvgain{1.19}} & \wvscore{95.50}{\wvgain{1.52}} & \wvscore{93.97}{\wvgain{1.02}} & \wvscore{92.20}{\wvgain{2.07}} \\
        \bottomrule
    \end{tabularx}%
    }
\vspace{-2mm}
\end{table}

Table~\ref{tab:web_generation_plugin_results} shows broad plug-and-play gains
across Web Generators. Across all seven models, WebViSE improves most
reported scores and achieves the highest result in six of the seven metrics.
GPT-5.6-Sol and Qwen3.8-27B improve on every metric, while Claude Opus 5 gains
3.25\% in WebGen-Bench Accuracy and 2.17 points in MiniAppBench Dynamic
despite its strong baseline. However, Kimi
K3, which has strong native frontend generation and visual reasoning
capabilities~\citep{kimiteam2026kimik3}, gains 1.41 points in Functionality and
0.89 points in Total, but decreases by 1.60 points in Aesthetics. This suggests
that additional Skill guidance may sometimes introduce design cues that compete
with a model's native style and reduce global visual coherence. Qwen3.8-Max and
DeepSeek-V4.1-Flash similarly show small decreases in Static but gains in
Dynamic. Because retrieved Visual Skills often encode interaction mechanisms
and state transitions, WebViSE may benefit interactive behavior more
consistently than static presentation.
A blinded pairwise human evaluation also prefers WebViSE over Base for every
generator, with margins of 12--34 percentage points; Appendix~\ref{app:human_protocol}
reports the protocol and full results.

\vspace{-2mm}
\subsubsection{Ablation Study: Effectiveness of WebViSE Guidance}
\label{sec:generation_guidance_ablation}
\vspace{-1mm}

We compare No Skill, General Design Skill, Random Skill, and
WebViSE using Claude Opus 5 and Qwen3.8-27B, with results in Table ~\ref{tab:web_generation_guidance_ablation}. The General Design Skill
baseline uses UI/UX Pro Max~\citep{nextlevelbuilder2026uiuxpromax} (over 125K
GitHub stars as of September 4, 2026). Random Skill provides four randomly
selected Visual Skills, matching the average number retrieved by WebViSE.

\begin{table}[h]
    \centering
    \caption{\textbf{Ablation of plug-and-play Skill guidance.} General Skill means  General Design Skill. Bold and underline denote the best and second-best
    scores in each model block.}
    \label{tab:web_generation_guidance_ablation}
    \footnotesize
    \setlength{\tabcolsep}{0.8pt}
    \renewcommand{\arraystretch}{1.04}
    \newcommand{\wvgain}[1]{\raisebox{0.15ex}{\scalebox{0.60}{\textcolor{green!45!black}{$\uparrow$#1}}}}
    \newcommand{\wvloss}[1]{\raisebox{0.15ex}{\scalebox{0.60}{\textcolor{red!75!black}{$\downarrow$#1}}}}
    \newcommand{\basevalue}[1]{\makebox[3.45em][l]{\makebox[2.20em][r]{#1}}}
    \newcommand{\wvscore}[2]{\makebox[3.45em][l]{\makebox[2.20em][r]{#1}\makebox[1.25em][l]{#2}}}
    \resizebox{0.98\linewidth}{!}{%
    \begin{tabularx}{\linewidth}{@{}L{0.15\linewidth}@{\hspace{1pt}}L{0.21\linewidth}@{\hspace{-3pt}}C{0.13\linewidth}@{\hspace{2pt}}*{6}{Y}@{}}
        \toprule
        \multirow{2}{=}{Model} & \multirow{2}{*}{Setting}
        & \multicolumn{1}{c}{WebGen-Bench}
        & \multicolumn{3}{c}{MiniAppBench}
        & \multicolumn{3}{c}{Cookie-Bench} \\
        \cmidrule(lr){3-3}\cmidrule(lr){4-6}\cmidrule(lr){7-9}
        & & Accuracy (\%) & Int. & Stat.& Dyn.
        & Func. & Aes. & Total \\
        \midrule
        \multirow{4}{=}{Claude Opus 5\\~\citep{anthropic2026opus5}}
        & No Skill & \basevalue{56.80} & \basevalue{\textbf{95.91}} & \basevalue{\underline{85.94}} & \basevalue{\underline{78.23}} & \basevalue{\textbf{97.81}} & \basevalue{95.46} & \basevalue{\underline{94.87}} \\
        & \textsc{+General Skill} & \wvscore{\underline{59.20}}{\wvgain{2.40}} & \wvscore{95.14}{\wvloss{0.77}} & \wvscore{85.88}{\wvloss{0.06}} & \wvscore{78.22}{\wvloss{0.01}} & \wvscore{96.04}{\wvloss{1.77}} & \wvscore{\underline{95.98}}{\wvgain{0.52}} & \wvscore{94.15}{\wvloss{0.72}} \\
        & \textsc{+Random Skill} & \wvscore{50.70}{\wvloss{6.10}} & \wvscore{90.31}{\wvloss{5.60}} & \wvscore{70.48}{\wvloss{15.46}} & \wvscore{64.93}{\wvloss{13.30}} & \wvscore{91.77}{\wvloss{6.04}} & \wvscore{92.43}{\wvloss{3.03}} & \wvscore{91.96}{\wvloss{2.91}} \\
        & \textsc{+WebViSE}  & \wvscore{\textbf{60.05}}{\wvgain{3.25}} & \wvscore{\underline{95.70}}{\wvloss{0.21}} & \wvscore{\textbf{86.44}}{\wvgain{0.50}} & \wvscore{\textbf{80.40}}{\wvgain{2.17}} & \wvscore{\underline{96.74}}{\wvloss{1.07}} & \wvscore{\textbf{96.52}}{\wvgain{1.06}} & \wvscore{\textbf{94.99}}{\wvgain{0.12}} \\
        \midrule
        \multirow{4}{=}{Qwen3.8-27B\\~\citep{qwen2026qwen3827b}}
        & No Skill & \basevalue{52.86} & \basevalue{\underline{93.45}} & \basevalue{\underline{79.06}} & \basevalue{68.85} & \basevalue{\underline{93.98}} & \basevalue{92.95} & \basevalue{\underline{90.13}} \\
        & \textsc{+General Skill} & \wvscore{\textbf{56.21}}{\wvgain{3.35}} & \wvscore{92.98}{\wvloss{0.47}} & \wvscore{78.55}{\wvloss{0.51}} & \wvscore{\underline{69.72}}{\wvgain{0.87}} & \wvscore{92.95}{\wvloss{1.03}} & \wvscore{91.72}{\wvloss{1.23}} & \wvscore{89.13}{\wvloss{1.00}} \\
        & \textsc{+Random Skill} & \wvscore{50.08}{\wvloss{2.78}} & \wvscore{85.92}{\wvloss{7.53}} & \wvscore{64.51}{\wvloss{14.55}} & \wvscore{56.91}{\wvloss{11.94}} & \wvscore{91.25}{\wvloss{2.73}} & \wvscore{\underline{93.35}}{\wvgain{0.40}} & \wvscore{88.99}{\wvloss{1.14}} \\
        & \textsc{+WebViSE} & \wvscore{\underline{54.40}}{\wvgain{1.54}} & \wvscore{\textbf{94.93}}{\wvgain{1.48}} & \wvscore{\textbf{80.63}}{\wvgain{1.57}} & \wvscore{\textbf{70.04}}{\wvgain{1.19}} & \wvscore{\textbf{95.50}}{\wvgain{1.52}} & \wvscore{\textbf{93.97}}{\wvgain{1.02}} & \wvscore{\textbf{92.20}}{\wvgain{2.07}} \\
        \bottomrule
    \end{tabularx}%
    }
    \vspace{-0.5\baselineskip}
\end{table}

\mypara{The value of Skill guidance depends on the generator's implementation capacity.}
For Qwen3.8-27B, General Design Skill produces the largest WebGen-Bench gain (+3.35\% versus +1.54\% with WebViSE), suggesting that broad guidance helps cover basic executable UI requirements. WebViSE nevertheless improves every reported metric, including Cookie-Bench Total by 2.07 points, but its smaller WebGen-Bench gain suggests that specialized mechanisms are harder for the smaller model to implement fully. For Claude Opus 5, General Design Skill improves only WebGen-Bench Accuracy (+2.40\%) and Cookie-Bench Aesthetics (+0.52). WebViSE instead improves Accuracy by 3.25 points, MiniAppBench Dynamic by 2.17 points, and Cookie-Bench Aesthetics by 1.06 points, while outperforming General Design Skill on every metric. Thus, the smaller model benefits more from broad guidance, whereas the stronger model can better realize specialized Visual Skills.

\mypara{Mismatched Skills degrade performance.}
Random Skill lowers WebGen-Bench Accuracy by 6.10\% for Claude Opus 5 and 2.78 points for Qwen3.8-27B. Cookie-Bench Total drops by 2.91 and 1.14 points, respectively. On MiniAppBench, Static drops by 15.46 and 14.55 points, while Dynamic drops by 13.30 and 11.94 points. Its only positive change is a 0.40-point Aesthetics gain for Qwen3.8-27B. These results show that mismatched guidance can substantially disrupt task realization, confirming the importance of retrieving Visual Skills that match the user instruction.
\vspace{-2mm}
\subsubsection{Case Study}
\label{sec:generation_case_study}
\vspace{-2mm}
\begin{figure*}[!t]
    \centering
    \includegraphics[width=0.88\textwidth]{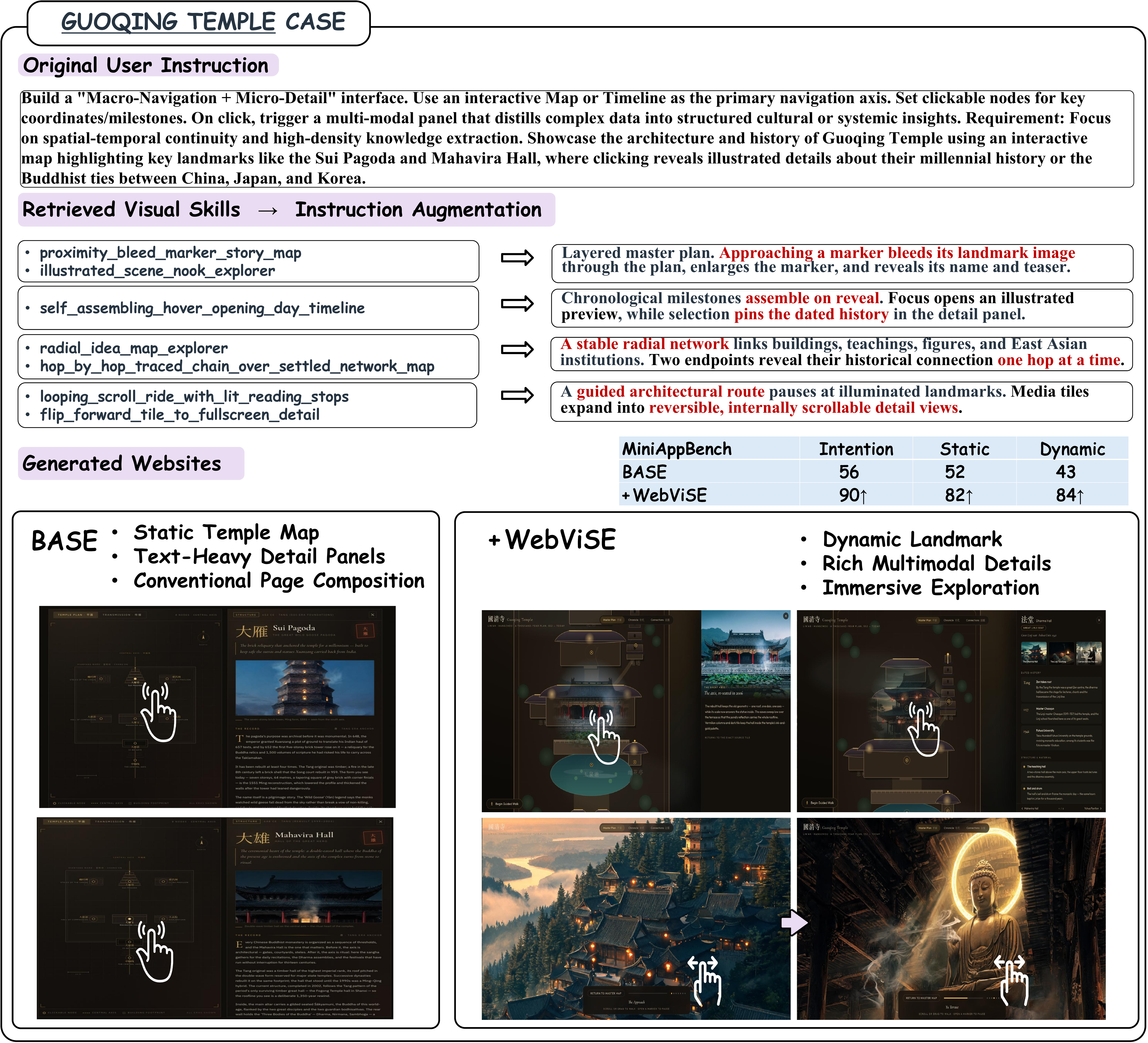}
    \vspace{-1\baselineskip}
\caption{\textbf{Guoqing Temple case study on MiniAppBench using Qwen3.8-27B.}
WebViSE retrieves Visual Skills and adds visual requirements to the Instruction for website generation. \textbf{For the full
interactive experience, we encourage readers to explore the live
\href{https://yifeisun-0710.github.io/WebViSE/case_study/original/}{\underline{Base}}
and
\href{https://yifeisun-0710.github.io/WebViSE/case_study/skill-expanded/}{\underline{WebViSE}}
websites.}}
\vspace{-1.5\baselineskip}
    \label{fig:guoqing_temple_case_study}
\end{figure*}

Figure~\ref{fig:guoqing_temple_case_study} compares two websites generated
from the same instruction. Base presents a mostly static temple map and
text-heavy panels. WebViSE adds landmark stories revealed near map markers,
interactive chronological milestones, a radial map of historical and religious
relationships, and an architectural route with expandable multimedia details.
Its Intention, Static, and Dynamic scores are 90/82/84, compared with
56/52/43 for Base.
For the full interactive experience, we encourage readers to explore the live \href{https://yifeisun-0710.github.io/WebViSE/case_study/original/}{\underline{Base}} and \href{https://yifeisun-0710.github.io/WebViSE/case_study/skill-expanded/}{\underline{WebViSE}} websites.

\vspace{-2mm}
\subsection{Internalizing Visual Design Intelligence}
\label{subsec:capability_internalization}
\vspace{-2mm}
\begin{table}[h]
    \centering
    \caption{\textbf{WebViSE internalization results for Qwen3.8-27B.}
    Bold denotes the best score per column; underline denotes the second best.}    \label{tab:web_generation_internalization}
    \footnotesize
    \setlength{\tabcolsep}{0.8pt}
    \renewcommand{\arraystretch}{1.04}
    \begin{tabularx}{\linewidth}{@{}L{0.28\linewidth}@{\hspace{4pt}}C{0.13\linewidth}@{\hspace{4pt}}*{6}{Y}@{}}
        \toprule
        \multirow{2}{=}{Method}
        & \multicolumn{1}{c}{WebGen-Bench}
        & \multicolumn{3}{c}{MiniAppBench}
        & \multicolumn{3}{c}{Cookie-Bench} \\
        \cmidrule(lr){2-2}\cmidrule(lr){3-5}\cmidrule(lr){6-8}
        & Accuracy (\%) & Int. & Stat. & Dyn.
        & Func. & Aes. & Total \\
        \midrule
        Qwen3.8-27B
        & 52.86 & 93.45 & 79.06 & 68.85 & 93.98 & 92.95 & 90.13 \\
        Qwen3.8-27B \textsc{+WebViSE}
        & \textbf{54.40} & \textbf{94.93} & \textbf{80.63} & \underline{70.04} & \textbf{95.50} & \underline{93.97} & \underline{92.20} \\
        Qwen3.8-27B NoSkill-SFT
        & \underline{54.33} & 92.70 & 79.25 & 68.10 & \underline{94.05} & 92.61 & 91.67\\
        Qwen3.8-27B WebViSE-SFT
        & 53.55 & \underline{93.61} & \underline{79.59} & \textbf{70.12} & 93.50 & \textbf{94.78} & \textbf{92.64} \\
        \bottomrule
    \end{tabularx}
    \vspace{-1\baselineskip}
\end{table}

To evaluate internalization, we compare Qwen3.8-27B~\citep{qwen2026qwen3827b},
the \textsc{+WebViSE} setting, NoSkill-SFT, and WebViSE-SFT. Both SFT variants use the
same 10,000 naturalistic instructions and internal WebDev teacher. NoSkill-SFT
generates training implementations from the original instructions, whereas
WebViSE-SFT augments them with retrieved Visual Skills. For student training, each
implementation is paired with its original instruction. At inference, both variants receive only the original instruction. Appendix~\ref{app:sft_training_configuration} reports the training configuration.

\vspace{-1mm}
\mypara{Skill guidance improves visual and dynamic quality after SFT.}
As shown in Table~\ref{tab:web_generation_internalization}, WebViSE-SFT
improves over the base model without inference-time Skill retrieval and
achieves the best MiniAppBench Dynamic and Cookie-Bench Aesthetics and Total
scores. Compared with the matched NoSkill-SFT control, it gains 2.02 points
on Dynamic, 2.17 on Aesthetics, and 0.97 on Total. However, WebGen-Bench
Accuracy falls by 0.78 percentage points and Cookie-Bench Functionality
by 0.55 points. Skill guidance thus helps visual quality and dynamic
interaction, but its benefits do not consistently extend to functional
correctness. Appendix~\ref{app:pointer_lab_internalization_case_study}
provides a case study.

\vspace{-2mm}
\section{Scaling Visual Skill Retrieval}
\label{sec:skill_retrieval_evaluation}
\vspace{-2mm}
\subsection{Experimental Setup}
\label{sec:retrieval_setup}

\mypara{Benchmarks.}
The \textbf{Internal benchmark} pairs 850 website-derived Goal Queries with 864
source-Skill assignments from the paired-view construction in
Section~\ref{subsec:skill_construction}. The \textbf{External benchmark}
contains 119 independently collected website-building intents with pooled
graded Skill-utility labels. Appendix~\ref{app:retrieval_protocol} details their construction and labeling protocols.

\mypara{Retrieval settings.}
We evaluate Flat LLM, the Initial DAG, and the Evolved DAG using Claude Opus 5
and Qwen3.8-Max as retrieval models. The methods start from the same source
Skill collection and use the same evaluation queries. When evolution merges
Skills, the resulting active Skill retains the identities of its constituent
source Skills. A retrieved active Skill occupies one rank and counts as
recovering only those constituent identities that are positives for the
current query. We report retrieval quality, average input and output tokens,
and estimated cost relative to Flat LLM within each retrieval model.
Appendix~\ref{app:retrieval_protocol} provides the full retrieval protocol.

\vspace{-2mm}
\subsection{Results}
\label{sec:retrieval_results}
 \vspace{-2mm}

Table~\ref{tab:retrieval_full_metrics} compares the DAGs with Flat LLM,
which processes the full Skill catalog for every request. With Claude Opus 5,
whose Flat LLM retrieval is already strong, the Evolved DAG maintains similar
MRR@10 (0.6061 versus 0.6028) and improves nDCG@10 (0.6198 versus 0.5971)
at 70.77\% of the estimated cost, although R@10 decreases (0.7762 versus
0.8079). With Qwen3.8-Max, whose Flat LLM scores are lower, the Evolved DAG
improves all seven quality metrics at 39.09\% of the estimated cost. The
Initial DAG provides most of the efficiency gain, while query-induced
evolution further improves the quality--cost tradeoff. Selective routing
avoids inspecting the full catalog for each query, motivating its use as
the Skill library grows. Appendix~\ref{app:additional_evaluation} provides
additional baselines, evolution ablations, and structural analyses.
\begin{table}[H]
    \centering
    \vspace{-1\baselineskip}
    \caption{\textbf{Retrieval quality and cost for the 864-Skill library.}
Flat LLM examines the full Skill catalog for each query; relative cost is
normalized to Flat LLM within each retrieval model.}
    \label{tab:retrieval_full_metrics}
    \scriptsize
    \setlength{\tabcolsep}{1.8pt}
    \renewcommand{\arraystretch}{1.04}
    \resizebox{\linewidth}{!}{%
    \begin{tabular}{@{}L{0.13\linewidth}L{0.12\linewidth}rrrrrrrcr@{}}
        \toprule
        \multirow{2}{=}{Model} & \multirow{2}{=}{Library structure}
        & \multicolumn{4}{c}{Internal source recovery}
        & \multicolumn{3}{c}{External Skill utility}
        & \multirow{2}{*}{Avg. input / output}
        & \multirow{2}{*}{Relative cost} \\
        \cmidrule(lr){3-6}\cmidrule(lr){7-9}
        & & R@1 & R@5 & R@10 & MRR@10
        & nDCG@5 & nDCG@10 & Utility R@10 & & \\
        \midrule
        \multirow{3}{=}{Claude Opus 5\\~\citep{anthropic2026opus5}}
        & Flat LLM & 0.5065 & \textbf{0.7268} & \textbf{0.8079} & 0.6028
        & 0.6441 & 0.5971 & 0.3437 & 118,822 / 2,975 & 100.00\% \\
        & Initial DAG & 0.5135 & 0.7056 & 0.7679 & 0.5974
        & 0.6388 & 0.6029 & 0.3422 & 77,531 / 3,549 & 71.26\% \\
        & Evolved DAG & \textbf{0.5229} & 0.7103 & 0.7762 & \textbf{0.6061}
        & \textbf{0.6675} & \textbf{0.6198} & \textbf{0.3523}
        & 74,774 / 3,968 & \textbf{70.77\%} \\
        \midrule
        \multirow{3}{=}{Qwen3.8-Max\\~\citep{qwen2026qwen38max}}
        & Flat LLM & 0.3756 & 0.6015 & 0.6879 & 0.4753
        & 0.3330 & 0.3255 & 0.1863 & 67,023 / 749 & 100.00\% \\
        & Initial DAG & 0.4129 & 0.6032 & 0.6785 & 0.4987
        & 0.4728 & 0.4419 & 0.2579 & 30,222 / 831 & 47.23\% \\
        & Evolved DAG & \textbf{0.4806} & \textbf{0.6671} & \textbf{0.7306}
        & \textbf{0.5637} & \textbf{0.5483} & \textbf{0.5095}
        & \textbf{0.2895} & 22,472 / 1,535 & \textbf{39.09\%} \\
        \bottomrule
    \end{tabular}%
    }
    \vspace{-1\baselineskip}
\end{table}

\section{Related Work}
\label{sec:related_work}
\vspace{-2mm}
LLM-based systems increasingly generate websites from natural-language or visual specifications, motivating benchmarks that evaluate visual fidelity, functionality, and interactive behavior \citep{si2025design2code,xiao2025interaction2code,lu2025webgenbench,zhang2026miniappbench,yang2026cookiebench}. Beyond evaluation, recent work improves visual design and webpage generation using real-world data that pairs screenshots with code in WebCode2M \citep{gui2025webcode2m}, explicit layout information in LayoutCoder \citep{wu2025layoutcoder}, modular grounding and planning in ScreenCoder \citep{jiang2025screencoder}, and spatial reasoning in LaySPA \citep{li2026layspa}. WebGen-Agent \citep{lu2025webgenagent} and WebGen-R1 \citep{jiang2026webgenr1} further improve end-to-end website generation by incorporating execution feedback and reinforcement learning. While these works focus primarily on generating or evaluating complete webpages, WebViSE instead extracts reusable visual design intelligence from implemented websites and transfers it to new generation tasks.

\vspace{-1.5mm}
A complementary line of work studies reusable Skills for LLM agents. CRAFT \citep{yuan2024craft} provides an early related formulation that creates and retrieves specialized toolsets, while more recent approaches construct Skills from external knowledge or agent experience, and evolve them automatically \citep{pan2026anything2skill,alzubi2026evoskill,zhang2026coevoskills,li2026codeskill}. As Skill libraries grow, SkillRet \citep{kang2026skillret}, SkillRouter \citep{zheng2026skillrouter}, Graph-of-Skills \citep{liu2026graphofskills}, and SkillDAG \citep{bai2026skilldag} study scalable and structured ways to select relevant Skills, while SKILL0 \citep{lu2026skill0} explores internalizing Skill knowledge into model parameters. WebViSE extends this line to visual design intelligence by jointly evolving Skills and a multi-parent routing DAG, enabling their use through plug-and-play guidance or Skill-augmented training.

\vspace{-1.5mm}
A comprehensive discussion is provided in Appendix~\ref{app:related_work}.

\section{Conclusion}
\label{sec:conclusion}
\vspace{-2mm}

We introduced WebViSE, a framework for extracting transferable visual design intelligence from well-designed websites on the open web. Across seven Website Generators and three complementary benchmarks, plug-and-play WebViSE improves most reported metrics, with particularly consistent gains in dynamic interaction. Controlled ablations show that retrieving applicable Skills is important for these gains, while blinded human evaluators favor WebViSE across all seven evaluated generators. Skill-augmented SFT further shows that this design intelligence can be internalized into model weights. For scalable Skill access, the evolved multi-parent DAG improves all retrieval-quality metrics with Qwen3.8-Max while requiring only 39.09\% of the retrieval cost of Flat LLM. Together, these results establish an automated path for extracting and organizing visual design intelligence from the open web and transferring it across diverse models for website generation, with structured retrieval supporting efficient Skill access as the library grows.
\subsection*{AI use statement}
In this work, we used generative AI tools for synthetic data generation and
to assist with research methodology, experimental design, method implementation,
and related research workflows. Generative AI models were also used as components
of the proposed framework, including Visual Skill extraction and refinement,
Potential Query generation, and other model-assisted stages described in the paper.
We did not use generative AI tools for formulating or proving mathematical claims,
and other required-disclosure tasks not relevant to this work are not applicable.

Additionally, we used generative AI tools to assist with literature retrieval and
summarization, software development and debugging, drafting and editing parts of
the manuscript, figure and artifact preparation, and improving writing clarity.
All AI-generated data and intermediate research artifacts used in our experiments
were subject to the filtering, validation, and evaluation procedures described in
the paper. AI-assisted code and experimental implementations were reviewed and
tested by the authors, and all manuscript text, claims, and reported results were
reviewed and verified by the authors.

We take responsibility for the final content of this work, including text, claims,
code, experimental results, and artifacts produced with the aid of generative AI.

\subsection*{Ethics statement}

Our study uses publicly accessible website implementations and blinded human
evaluation. Appendix~\ref{app:setup_details} documents the source collections
and their applicable terms, and Appendix~\ref{app:human_protocol} describes the
pairwise evaluation protocol, including output anonymization, randomized
presentation order, and evaluation criteria. Source materials remain subject to
their original licenses and terms.

\subsection*{Reproducibility statement}

We release the implementation, extracted Visual Skills, and associated Skill DAG
in our \href{https://github.com/YifeiSun-0710/WebViSE}{code
repository}. Appendix~\ref{app:setup_details} reports the data construction,
optimization, training, benchmark, and human-evaluation settings;
Appendix~\ref{app:additional_evaluation} provides complete retrieval protocols
and additional analyses; and Appendix~\ref{app:model_facing_prompts} provides
the model-facing prompts.

\bibliographystyle{iclr2027_conference}
\bibliography{iclr2027_conference}

\begin{thebibliography}{46}
\providecommand{\natexlab}[1]{#1}
\providecommand{\url}[1]{\texttt{#1}}
\expandafter\ifx\csname urlstyle\endcsname\relax
  \providecommand{\doi}[1]{doi: #1}\else
  \providecommand{\doi}{doi: \begingroup \urlstyle{rm}\Url}\fi

\bibitem[Alzubi et~al.(2026)Alzubi, Provenzano, Bingham, Chen, and Vu]{alzubi2026evoskill}
Salaheddin Alzubi, Noah Provenzano, Jaydon Bingham, Weiyuan Chen, and Tu~Vu.
\newblock {EvoSkill}: Automated skill discovery for multi-agent systems.
\newblock \emph{arXiv preprint arXiv:2603.02766}, 2026.

\bibitem[{Anthropic}(2025)]{anthropic2025agentskills}
{Anthropic}.
\newblock Equipping agents for the real world with agent skills.
\newblock Anthropic Engineering, 2025.
\newblock URL \url{https://www.anthropic.com/engineering/equipping-agents-for-the-real-world-with-agent-skills}.

\bibitem[{Anthropic}(2026)]{anthropic2026opus5}
{Anthropic}.
\newblock Introducing {Claude Opus 5}.
\newblock Anthropic News, 2026.
\newblock URL \url{https://www.anthropic.com/news/claude-opus-5}.

\bibitem[Bai et~al.(2026)Bai, Wan, Zhou, Yu, You, and Tsang]{bai2026skilldag}
Tong Bai, Zhenglin Wan, Pengfei Zhou, Xingrui Yu, Yang You, and Ivor~W. Tsang.
\newblock {SkillDAG}: Self-evolving typed skill graphs for {LLM} skill selection at scale.
\newblock \emph{arXiv preprint arXiv:2606.03056}, 2026.

\bibitem[{ByteDance Seed}(2026)]{bytedanceseed2026seed20}
{ByteDance Seed}.
\newblock Seed2.0 model card: Towards intelligence frontier for real-world complexity.
\newblock \emph{arXiv preprint arXiv:2607.00248}, 2026.

\bibitem[Cormack et~al.(2009)Cormack, Clarke, and B{\"u}ttcher]{cormack2009rrf}
Gordon~V. Cormack, Charles L.~A. Clarke, and Stefan B{\"u}ttcher.
\newblock Reciprocal rank fusion outperforms condorcet and individual rank learning methods.
\newblock In \emph{International ACM SIGIR Conference on Research and Development in Information Retrieval (SIGIR)}, 2009.

\bibitem[Dai et~al.(2026)Dai, Wu, Fang, Li, and Wang]{dai2026iwebgenbench}
Dasen Dai, Biao Wu, Meng Fang, Shuoqi Li, and Wenhao Wang.
\newblock {I-WebGenBench}: Evaluating interactivity in {LLM}-generated scientific web applications.
\newblock \emph{arXiv preprint arXiv:2606.00750}, 2026.

\bibitem[{DeepSeek-AI}(2026)]{deepseekai2026deepseekv41flash}
{DeepSeek-AI}.
\newblock {DeepSeek-V4.1-Flash} release.
\newblock DeepSeek API Documentation, 2026.
\newblock URL \url{https://api-docs.deepseek.com/updates/}.

\bibitem[Gui et~al.(2025)Gui, Li, Wan, Shi, Zhang, Su, Chen, Chen, Wu, Zhou, Jiang, Jin, and Zhang]{gui2025webcode2m}
Yi~Gui, Zhen Li, Yao Wan, Yemin Shi, Hongyu Zhang, Yi~Su, Bohua Chen, Dongping Chen, Siyuan Wu, Xing Zhou, Wenbin Jiang, Hai Jin, and Xiangliang Zhang.
\newblock {WebCode2M}: A real-world dataset for code generation from webpage designs.
\newblock In \emph{The ACM Web Conference (WWW)}, 2025.

\bibitem[Gupta et~al.(2025)Gupta, Chang, Bui, Hsieh, and Dhillon]{gupta2025lattice}
Nilesh Gupta, Wei-Cheng Chang, Ngot Bui, Cho-Jui Hsieh, and Inderjit~S. Dhillon.
\newblock {LLM}-guided hierarchical search for end-to-end reasoning intensive retrieval.
\newblock \emph{arXiv preprint arXiv:2510.13217}, 2025.

\bibitem[Hu et~al.(2026)Hu, Li, Ramakrishnan, and Zhao]{hu2026erm}
Yuntong Hu, Sha Li, Naren Ramakrishnan, and Liang Zhao.
\newblock {RAG} without forgetting: Continual query-infused key memory.
\newblock In \emph{International Conference on Machine Learning (ICML)}, 2026.

\bibitem[{ISO}(2018)]{iso9241_11_2018}
{ISO}.
\newblock Ergonomics of human-system interaction---part 11: Usability: Definitions and concepts.
\newblock International Standard ISO 9241-11:2018, International Organization for Standardization, 2018.

\bibitem[{ISO}(2019)]{iso9241_210_2019}
{ISO}.
\newblock Ergonomics of human-system interaction---part 210: Human-centred design for interactive systems.
\newblock International Standard ISO 9241-210:2019, International Organization for Standardization, 2019.

\bibitem[{ISO}(2020)]{iso9241_110_2020}
{ISO}.
\newblock Ergonomics of human-system interaction---part 110: Interaction principles.
\newblock International Standard ISO 9241-110:2020, International Organization for Standardization, 2020.

\bibitem[Jiang et~al.(2026)Jiang, Cai, Park, Shen, Kim, Li, and Wang]{jiang2026webgenr1}
Juyong Jiang, Chenglin Cai, Chansung Park, Jiasi Shen, Sunghun Kim, Jianguo Li, and Yue Wang.
\newblock {WebGen-R1}: Incentivizing large language models to generate functional and aesthetic websites with reinforcement learning.
\newblock \emph{arXiv preprint arXiv:2604.20398}, 2026.

\bibitem[Jiang et~al.(2025)Jiang, Zheng, Wan, Han, Wang, Lyu, and Yue]{jiang2025screencoder}
Yilei Jiang, Yaozhi Zheng, Yuxuan Wan, Jiaming Han, Qunzhong Wang, Michael~R. Lyu, and Xiangyu Yue.
\newblock {ScreenCoder}: Advancing visual-to-code generation for front-end automation via modular multimodal agents.
\newblock \emph{arXiv preprint arXiv:2507.22827}, 2025.

\bibitem[Kang et~al.(2025)Kang, Li, Liu, Chen, Zhang, Jiang, Yu, and Su]{kang2025pqr}
Junfeng Kang, Rui Li, Qi~Liu, Yanjiang Chen, Zheng Zhang, Junzhe Jiang, Heng Yu, and Yu~Su.
\newblock {PQR}: Improving dense retrieval via potential query modeling.
\newblock In \emph{Annual Meeting of the Association for Computational Linguistics (ACL)}, 2025.

\bibitem[Kang et~al.(2026)Kang, Cho, and Kim]{kang2026skillret}
Ryangkyung Kang, Hongcheol Cho, and Youngeun Kim.
\newblock {SkillRet}: A large-scale benchmark for skill retrieval in {LLM} agents.
\newblock \emph{arXiv preprint arXiv:2605.05726}, 2026.

\bibitem[{Kimi Team}(2026)]{kimiteam2026kimik3}
{Kimi Team}.
\newblock {Kimi K3}: Open frontier intelligence.
\newblock \emph{arXiv preprint arXiv:2607.24653}, 2026.

\bibitem[Li et~al.(2026{\natexlab{a}})Li, Petrangeli, Shen, and Chen]{li2026layspa}
Sha Li, Stefano Petrangeli, Yu~Shen, and Xiang Chen.
\newblock From pixels to policies: Reinforcing spatial reasoning in language models for content-aware layout design.
\newblock \emph{arXiv preprint arXiv:2602.13912}, 2026{\natexlab{a}}.

\bibitem[Li et~al.(2026{\natexlab{b}})Li, Zhang, Zhang, Liu, and Liu]{li2026codeskill}
Yanzhou Li, Yiran Zhang, Xiaoyu Zhang, Xiaoxia Liu, and Yang Liu.
\newblock {CODESKILL}: Learning self-evolving skills for coding agents.
\newblock \emph{arXiv preprint arXiv:2605.25430}, 2026{\natexlab{b}}.

\bibitem[Liang et~al.(2026)Liang, Wang, Liang, and Liu]{liang2026skillstructure}
Qiliang Liang, Hansi Wang, Zhong Liang, and Yang Liu.
\newblock From skill text to skill structure: The scheduling-structural-logical representation for agent skills.
\newblock \emph{arXiv preprint arXiv:2604.24026}, 2026.

\bibitem[Liu et~al.(2026)Liu, Li, Du, Wu, Gui, Kuang, and Sun]{liu2026graphofskills}
Dawei Liu, Zongxia Li, Hongyang Du, Xiyang Wu, Shihang Gui, Yongbei Kuang, and Lichao Sun.
\newblock {Graph-of-Skills}: Dependency-aware structural retrieval for massive agent skills.
\newblock \emph{arXiv preprint arXiv:2604.05333}, 2026.

\bibitem[Lu et~al.(2026)Lu, Yao, Wu, Han, Gu, Cai, Lu, Xiao, Zhuang, and Shen]{lu2026skill0}
Zhengxi Lu, Zhiyuan Yao, Jinyang Wu, Chengcheng Han, Qi~Gu, Xunliang Cai, Weiming Lu, Jun Xiao, Yueting Zhuang, and Yongliang Shen.
\newblock {SKILL0}: In-context agentic reinforcement learning for skill internalization.
\newblock \emph{arXiv preprint arXiv:2604.02268}, 2026.

\bibitem[Lu et~al.(2025{\natexlab{a}})Lu, Ren, Yang, Wang, Zong, Pan, Zhan, and Li]{lu2025webgenagent}
Zimu Lu, Houxing Ren, Yunqiao Yang, Ke~Wang, Zhuofan Zong, Junting Pan, Mingjie Zhan, and Hongsheng Li.
\newblock {WebGen-Agent}: Enhancing interactive website generation with multi-level feedback and step-level reinforcement learning.
\newblock \emph{arXiv preprint arXiv:2509.22644}, 2025{\natexlab{a}}.

\bibitem[Lu et~al.(2025{\natexlab{b}})Lu, Yang, Ren, Hou, Xiao, Wang, Shi, Zhou, Zhan, and Li]{lu2025webgenbench}
Zimu Lu, Yunqiao Yang, Houxing Ren, Haotian Hou, Han Xiao, Ke~Wang, Weikang Shi, Aojun Zhou, Mingjie Zhan, and Hongsheng Li.
\newblock {WebGen-Bench}: Evaluating {LLM}s on generating interactive and functional websites from scratch.
\newblock In \emph{Conference on Neural Information Processing Systems (NeurIPS)}, 2025{\natexlab{b}}.

\bibitem[Mi et~al.(2026)Mi, Ma, Yang, Li, Wang, Zhang, and Wang]{mi2026skillpro}
Qirui Mi, Zhijian Ma, Mengyue Yang, Haoxuan Li, Yisen Wang, Haifeng Zhang, and Jun Wang.
\newblock {Skill-Pro}: Learning reusable skills from experience via non-parametric {PPO} for {LLM} agents.
\newblock In \emph{International Conference on Machine Learning (ICML)}, 2026.

\bibitem[{nextlevelbuilder}(2026)]{nextlevelbuilder2026uiuxpromax}
{nextlevelbuilder}.
\newblock {UI/UX Pro Max}.
\newblock GitHub repository, 2026.
\newblock URL \url{https://github.com/nextlevelbuilder/ui-ux-pro-max-skill}.
\newblock Accessed September 4, 2026.

\bibitem[{OpenAI}(2025)]{openai2025gpt5}
{OpenAI}.
\newblock Introducing gpt-5.
\newblock OpenAI, 2025.
\newblock URL \url{https://openai.com/index/introducing-gpt-5/}.

\bibitem[{OpenAI}(2026{\natexlab{a}})]{openai2026gpt54}
{OpenAI}.
\newblock Introducing gpt-5.4.
\newblock OpenAI, 2026{\natexlab{a}}.
\newblock URL \url{https://openai.com/index/introducing-gpt-5-4/}.

\bibitem[{OpenAI}(2026{\natexlab{b}})]{openai2026gpt56}
{OpenAI}.
\newblock {GPT-5.6 Preview System Card}.
\newblock Technical report, OpenAI, July 2026{\natexlab{b}}.
\newblock URL \url{https://deploymentsafety.openai.com/gpt-5-6/gpt-5-6.pdf}.

\bibitem[Pan et~al.(2026)Pan, Yang, Li, Zhou, Chen, Li, Chen, and He]{pan2026anything2skill}
Qianjun Pan, Yutao Yang, Junsong Li, Jie Zhou, Kai Chen, Xin Li, Qin Chen, and Liang He.
\newblock {Anything2Skill}: Compiling external knowledge into reusable skills for agents.
\newblock \emph{arXiv preprint arXiv:2606.09316}, 2026.

\bibitem[{Qwen Team}(2026{\natexlab{a}})]{qwen2026qwen3827b}
{Qwen Team}.
\newblock {Qwen3.8-27B}.
\newblock Hugging Face, 2026{\natexlab{a}}.
\newblock URL \url{https://huggingface.co/Qwen/Qwen3.8-27B}.
\newblock Accessed September 23, 2026.

\bibitem[{Qwen Team}(2026{\natexlab{b}})]{qwen2026qwen38max}
{Qwen Team}.
\newblock {Qwen3.8-Max}: A new bar for coding and cowork.
\newblock Qwen Blog, 2026{\natexlab{b}}.
\newblock URL \url{https://qwen.ai/blog?id=qwen3.8}.

\bibitem[Robertson \& Walker(1994)Robertson and Walker]{robertson1994bm25}
Stephen~E. Robertson and Steve Walker.
\newblock Some simple effective approximations to the 2-poisson model for probabilistic weighted retrieval.
\newblock In \emph{International ACM SIGIR Conference on Research and Development in Information Retrieval (SIGIR)}, 1994.

\bibitem[Sarthi et~al.(2024)Sarthi, Abdullah, Tuli, Khanna, Goldie, and Manning]{sarthi2024raptor}
Parth Sarthi, Salman Abdullah, Aditi Tuli, Shubh Khanna, Anna Goldie, and Christopher~D. Manning.
\newblock {RAPTOR}: Recursive abstractive processing for tree-organized retrieval.
\newblock In \emph{International Conference on Learning Representations (ICLR)}, 2024.

\bibitem[Si et~al.(2025)Si, Zhang, Li, Yang, Liu, and Yang]{si2025design2code}
Chenglei Si, Yanzhe Zhang, Ryan Li, Zhengyuan Yang, Ruibo Liu, and Diyi Yang.
\newblock {Design2Code}: Benchmarking multimodal code generation for automated front-end engineering.
\newblock In \emph{Annual Conference of the Nations of the Americas Chapter of the Association for Computational Linguistics (NAACL)}, 2025.

\bibitem[Wu et~al.(2025)Wu, Gao, Li, Wen, and Liao]{wu2025layoutcoder}
Fan Wu, Cuiyun Gao, Shuqing Li, Xin-Cheng Wen, and Qing Liao.
\newblock {MLLM}-based {UI2Code} automation guided by {UI} layout information.
\newblock In \emph{ACM SIGSOFT International Symposium on Software Testing and Analysis (ISSTA)}, 2025.

\bibitem[Xiao et~al.(2025)Xiao, Wan, Huo, Wang, Xu, Wang, Xu, Wang, and Lyu]{xiao2025interaction2code}
Jingyu Xiao, Yuxuan Wan, Yintong Huo, Zixin Wang, Xinyi Xu, Wenxuan Wang, Zhiyao Xu, Yuhang Wang, and Michael~R. Lyu.
\newblock {Interaction2Code}: Benchmarking {MLLM}-based interactive webpage code generation from interactive prototyping.
\newblock In \emph{IEEE/ACM International Conference on Automated Software Engineering (ASE)}, 2025.

\bibitem[Yang et~al.(2026)Yang, Shen, Ding, Lou, Kou, Yu, Li, Wu, Bao, Liu, and Wu]{yang2026cookiebench}
Haoyue Yang, Zhangxiao Shen, Fan Ding, Hangting Lou, Yifeng Kou, Haoqing Yu, Jingyao Li, Zhengfan Wu, Siqi Bao, Jing Liu, and Hua Wu.
\newblock {Cookie-Bench}: Continuous on-screen key interaction evaluation for web generation.
\newblock \emph{arXiv preprint arXiv:2605.30000}, 2026.

\bibitem[Yuan et~al.(2024)Yuan, Chen, Wang, Fung, Peng, and Ji]{yuan2024craft}
Lifan Yuan, Yangyi Chen, Xingyao Wang, Yi~Fung, Hao Peng, and Heng Ji.
\newblock {CRAFT}: Customizing {LLM}s by creating and retrieving from specialized toolsets.
\newblock In \emph{International Conference on Learning Representations (ICLR)}, 2024.

\bibitem[{Z.ai}(2026)]{zhipuai2026glm53}
{Z.ai}.
\newblock {GLM-5.3}: Frontier coding with emergent cyber capabilities.
\newblock Z.ai Blog, 2026.
\newblock URL \url{https://z.ai/blog/glm-5.3}.

\bibitem[Zhang et~al.(2026{\natexlab{a}})Zhang, Fan, Zou, Chen, Wang, Zhou, Li, Huang, Yao, Zheng, Liu, Li, and Yu]{zhang2026coevoskills}
Hanrong Zhang, Shicheng Fan, Henry~Peng Zou, Yankai Chen, Zhenting Wang, Jiayu Zhou, Chengze Li, Wei-Chieh Huang, Yifei Yao, Kening Zheng, Xue Liu, Xiaoxiao Li, and Philip~S. Yu.
\newblock {CoEvoSkills}: Self-evolving agent skills via co-evolutionary verification.
\newblock \emph{arXiv preprint arXiv:2604.01687}, 2026{\natexlab{a}}.

\bibitem[Zhang et~al.(2025)Zhang, Li, Long, Zhang, Lin, Yang, Xie, Yang, Liu, Lin, Huang, and Zhou]{zhang2025qwen3embedding}
Yanzhao Zhang, Mingxin Li, Dingkun Long, Xin Zhang, Huan Lin, Baosong Yang, Pengjun Xie, An~Yang, Dayiheng Liu, Junyang Lin, Fei Huang, and Jingren Zhou.
\newblock {Qwen3 Embedding}: Advancing text embedding and reranking through foundation models.
\newblock \emph{arXiv preprint arXiv:2506.05176}, 2025.

\bibitem[Zhang et~al.(2026{\natexlab{b}})Zhang, Yu, Li, Zhuang, Mo, and Li]{zhang2026miniappbench}
Zuhao Zhang, Chengyue Yu, Yuante Li, Chenyi Zhuang, Linjian Mo, and Shuai Li.
\newblock {MiniAppBench}: Evaluating the shift from text to interactive {HTML} responses in {LLM}-powered assistants.
\newblock \emph{arXiv preprint arXiv:2603.09652}, 2026{\natexlab{b}}.

\bibitem[Zheng et~al.(2026)Zheng, Zhang, Ma, Yu, Zhu, Wu, Xu, Dong, Zhu, Huang, and Yu]{zheng2026skillrouter}
YanZhao Zheng, ZhenTao Zhang, Chao Ma, YuanQiang Yu, JiHuai Zhu, Yong Wu, Tianze Xu, Baohua Dong, Hangcheng Zhu, Ruohui Huang, and Gang Yu.
\newblock {SkillRouter}: Skill routing for {LLM} agents at scale.
\newblock \emph{arXiv preprint arXiv:2603.22455}, 2026.

\end{thebibliography}

\appendix
\section{Extensive Related Work}
\label{app:related_work}

\paragraph{Website Generation and Visual Design.}
Automated website generation spans reference-conditioned interface reconstruction and instruction-conditioned website creation. Design2Code evaluates the reconstruction of reference webpages from screenshots through visual elements, layout, text, position, and color \citep{si2025design2code}, while Interaction2Code extends this setting to dynamic interactions and identifies persistent difficulties with fine-grained state changes and complex transformations \citep{xiao2025interaction2code}. Instruction-conditioned benchmarks evaluate complementary aspects of complete website generation: WebGen-Bench measures functional correctness through executable operation--outcome tests \citep{lu2025webgenbench}; MiniAppBench evaluates intention alignment, static presentation, and dynamic interaction \citep{zhang2026miniappbench}; and Cookie-Bench combines static perception with autonomous interaction \citep{yang2026cookiebench}. Recent systems directly improve the generator itself. WebGen-Agent iteratively refines websites using screenshot and GUI-agent feedback and further trains the generator with step-level reinforcement learning \citep{lu2025webgenagent}, whereas WebGen-R1 couples scaffold-constrained generation with execution-grounded functional and visual rewards \citep{jiang2026webgenr1}. These methods reconstruct a supplied design, optimize an end-to-end generator, or evaluate its outputs.

Large-scale data and structured generation methods further improve visual design and webpage generation. WebCode2M renders and filters real-world web resources into 2.56 million paired design images, webpage code, and layout details \citep{gui2025webcode2m}. LayoutCoder extracts element relations and layout trees from webpage images to guide code generation and fusion \citep{wu2025layoutcoder}. ScreenCoder decomposes visual-to-code generation into grounding, hierarchical layout planning, and code generation, and uses the resulting pipeline as a scalable image--code data engine \citep{jiang2025screencoder}. LaySPA complements code-generation systems by training LLMs to reason over canvas geometry, element attributes, and spatial relations for content-aware graphic layout design \citep{li2026layspa}. These works provide data, reconstruction pipelines, or spatial policies for producing complete layouts. WebViSE instead extracts transferable visual and interaction mechanisms from well-designed websites and represents them as reusable Skills for new website generation tasks.

\paragraph{Reusable Agent Skills and Automated Skill Acquisition.}
Agent Skills package procedural knowledge, instructions, code, and supporting resources into modular capabilities that can be loaded on demand \citep{anthropic2025agentskills}. CRAFT derives specialized toolsets from successful solutions \citep{yuan2024craft}, Anything2Skill compiles heterogeneous external knowledge into structured Skills \citep{pan2026anything2skill}, and Skill-Pro learns reusable Skills from agent experience \citep{mi2026skillpro}. More recent work automates Skill construction and maintenance. EvoSkill diagnoses execution failures and proposes additions or revisions to structured Skill folders \citep{alzubi2026evoskill}; CoEvoSkills couples a Skill Generator with a co-evolving Surrogate Verifier to refine multi-file Skill packages \citep{zhang2026coevoskills}; and CODESKILL learns a management policy that extracts and maintains procedural Skills from coding-agent trajectories \citep{li2026codeskill}. These approaches primarily capture task-solving procedures from external documents, solution trajectories, or execution feedback. WebViSE instead derives reusable visual and interaction Skills from complementary intent and behavioral views of well-designed websites, while removing source-specific content and presentation.

\paragraph{Scalable Skill Retrieval and Structured Organization.}
As Skill libraries expand, selecting a small applicable subset becomes necessary for controlling context and avoiding irrelevant guidance. SkillRet studies retrieval-specific training over Skill collections \citep{kang2026skillret}, while SkillRouter combines metadata retrieval with full-document reranking \citep{zheng2026skillrouter}. Query-informed retrieval methods reduce mismatches between requests and indexed documents: PQR synthesizes potential queries to enrich document representations \citep{kang2025pqr}, whereas ERM incorporates successful historical query expansions into evolving retrieval keys \citep{hu2026erm}. Hierarchical methods such as RAPTOR and LATTICE provide coarse-to-fine corpus access through tree-structured representations \citep{sarthi2024raptor,gupta2025lattice}. Most closely related to WebViSE, Graph-of-Skills constructs an offline dependency graph and retrieves bounded Skill bundles through graph propagation \citep{liu2026graphofskills}, while SkillDAG maintains typed inter-Skill relations and adds execution-supported edges during online use \citep{bai2026skilldag}. WebViSE differs in both graph semantics and evolution supervision. Its multi-parent DAG organizes Visual Skills by their semantic design roles rather than executable dependencies, and evolves offline using Skill-conditioned Potential Queries as automatic retrieval supervision.

\paragraph{Skill Augmentation and Internalization.}
Most Agent Skill systems extend a frozen model by retrieving Skill content at inference time. This provides modularity but introduces retrieval errors and recurring context costs. SKILL0 instead internalizes procedural Skills through an agentic reinforcement-learning curriculum that begins with full Skill context and progressively withdraws it until the policy operates without runtime retrieval \citep{lu2026skill0}. WebViSE follows a complementary training path designed for website generation. Retrieved Visual Skills augment training instructions provided to a teacher Website Generator, and the resulting implementations are paired with the original instructions for supervised fine-tuning. Consequently, the target model receives neither Skill documents nor expanded instructions at inference time. This transfers visual design intelligence into model weights through Skill-guided data synthesis rather than progressive Skill-context removal.

\section{Qualitative Website Comparisons}
\label{app:opus5_motivation_cases}

Figure~\ref{fig:skill_usage} provides a detailed comparison on two sections of
one World Cup website. Table~\ref{tab:opus5_motivation_cases} presents the
complete collection of 16 controlled qualitative cases. Each row uses the same
original instruction and Claude Opus 5 generator across all three settings.
The second column reports all selected Website-Derived Visual Skills by canonical name.
The original instructions and interactive pages are available at
\url{https://yifeisun-0710.github.io/WebViSE/showcase/}.

\mypara{Qualitative observations.}
Across the collection, No Skill and the general website-design Skill often
retain familiar task templates, such as conventional dashboards, editors, and
centered hero layouts, even when the latter improves local polish.
Website-Derived Visual Skills instead introduce visibly broader design
languages and interaction structures, including orbitable 3D scenes, layered
cutaways, scroll-scrubbed narratives, pointer-responsive effects, and
simulation-specific motion. The increase in diversity is therefore not limited
to colors or decoration: it changes spatial composition, depth, motion, and
control affordances while keeping each result recognizable as the requested
application. Because each row fixes both the instruction and generator, these
contrasts isolate the effect of more specific mechanism-level visual guidance.

\begingroup
\setlength{\tabcolsep}{1pt}
\renewcommand{\arraystretch}{1.00}

\newcommand{\instructionpanel}[1]{%
  \includegraphics[width=\linewidth]{figure/motivation_cases/queries/case-#1.pdf}%
}
\newcommand{\resultpanel}[2]{%
  \includegraphics[width=\linewidth]{figure/motivation_cases/#1/#2.pdf}%
}
\newenvironment{skilllist}{%
  \fontsize{4.2}{4.5}\selectfont
  \begin{itemize}[leftmargin=0.9em,labelsep=0.25em,itemsep=0pt,parsep=0pt,
                  topsep=0pt,partopsep=0pt]%
}{%
  \end{itemize}%
}
\newcommand{\skillname}[1]{\item \nolinkurl{#1}}

\begin{longtable}{@{}>{\raggedright\arraybackslash}m{0.235\linewidth}
                    >{\raggedright\arraybackslash}m{0.145\linewidth}
                    >{\centering\arraybackslash}m{0.195\linewidth}
                    >{\centering\arraybackslash}m{0.195\linewidth}
                    >{\centering\arraybackslash}m{0.195\linewidth}@{}}
\caption{\textbf{Claude Opus 5 websites under three generation settings.} Each row fixes
the original instruction and generator while varying the visual guidance.}
\label{tab:opus5_motivation_cases}\\
\toprule
\textbf{Instruction} & \textbf{Selected Skill} & \textbf{No Skill} &
\textbf{General Website Design Skill} & \textbf{Ours: Website-Derived Visual Skill} \\
\midrule
\endfirsthead

\multicolumn{5}{c}{\tablename~\thetable\ continued}\\
\toprule
\textbf{Instruction} & \textbf{Selected Skill} & \textbf{No Skill} &
\textbf{General Website Design Skill} & \textbf{Ours: Website-Derived Visual Skill} \\
\midrule
\endhead

\midrule
\multicolumn{5}{r}{Continued on the next page}\\
\endfoot

\bottomrule
\endlastfoot

\instructionpanel{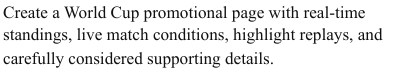} &
\begin{skilllist}
\skillname{split_flap_live_value_board}
\skillname{center_locked_media_row_over_live_backdrop}
\skillname{aperture_framed_scroll_scrubbed_clip}
\end{skilllist} &
\resultpanel{world-cup}{no_skill} &
\resultpanel{world-cup}{general_skill} &
\resultpanel{world-cup}{visual_skill} \\
\midrule

\instructionpanel{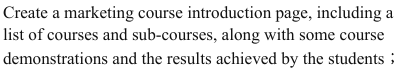} &
\begin{skilllist}
\skillname{center_out_photo_bloom_with_pointer_drift}
\skillname{edge_following_band_swaps_name_for_drifting_work_run}
\skillname{receding_stage_topic_slides}
\skillname{roaming_contact_sheet_with_lifted_detail}
\end{skilllist} &
\resultpanel{p-03}{no_skill} &
\resultpanel{p-03}{general_skill} &
\resultpanel{p-03}{visual_skill} \\
\midrule

\instructionpanel{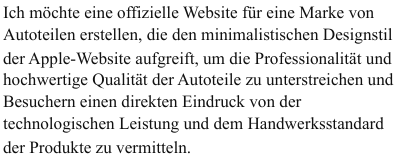} &
\begin{skilllist}
\skillname{scroll_scrubbed_object_opening_walkthrough}
\end{skilllist} &
\resultpanel{p-06}{no_skill} &
\resultpanel{p-06}{general_skill} &
\resultpanel{p-06}{visual_skill} \\
\midrule

\instructionpanel{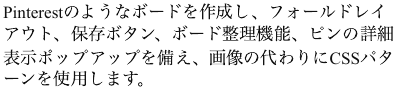} &
\begin{skilllist}
\skillname{flip_forward_tile_to_fullscreen_detail}
\skillname{spotlight_take_with_picture_trail_to_counter}
\skillname{tilting_stack_card_library}
\skillname{labelled_contrast_flipping_pointer_marker}
\skillname{pointer_trailing_soft_pattern_layer}
\skillname{word_as_window_texture_variant_stage}
\end{skilllist} &
\resultpanel{p-13}{no_skill} &
\resultpanel{p-13}{general_skill} &
\resultpanel{p-13}{visual_skill} \\
\midrule

\instructionpanel{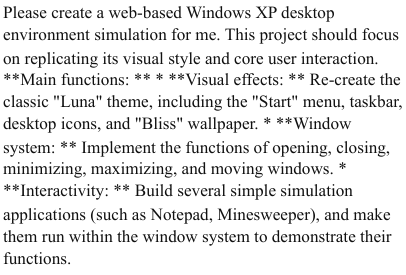} &
\begin{skilllist}
\skillname{booting_live_screen_stage}
\skillname{rippling_water_photo_backdrop}
\skillname{flat_fill_panel_curtain_lift_entry}
\end{skilllist} &
\resultpanel{p-16}{no_skill} &
\resultpanel{p-16}{general_skill} &
\resultpanel{p-16}{visual_skill} \\
\midrule

\instructionpanel{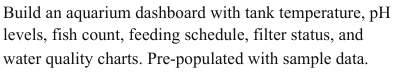} &
\begin{skilllist}
\skillname{water_shaped_period_row_with_in_column_time_scrub}
\skillname{charging_console_panel_choice_row}
\skillname{cursor_bent_drifting_particle_current}
\end{skilllist} &
\resultpanel{p-23}{no_skill} &
\resultpanel{p-23}{general_skill} &
\resultpanel{p-23}{visual_skill} \\
\midrule

\instructionpanel{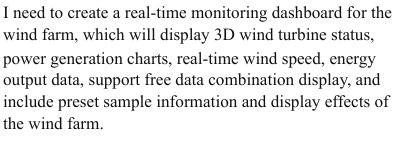} &
\begin{skilllist}
\skillname{water_shaped_period_row_with_in_column_time_scrub}
\skillname{operable_interface_on_orbitable_machine}
\skillname{neon_outline_trio_orbit_stage}
\skillname{charging_console_panel_choice_row}
\skillname{electric_arc_charge_button}
\skillname{cursor_bent_drifting_particle_current}
\end{skilllist} &
\resultpanel{p-24}{no_skill} &
\resultpanel{p-24}{general_skill} &
\resultpanel{p-24}{visual_skill} \\
\midrule

\instructionpanel{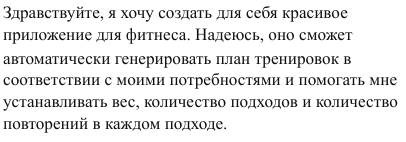} &
\begin{skilllist}
\skillname{graduated_rotary_dial_slot_filling}
\skillname{dot_matrix_level_plates}
\skillname{unit_mark_span_progress_with_forward_preview}
\skillname{spotlight_take_with_picture_trail_to_counter}
\skillname{sketched_mark_answer_sheet}
\skillname{charging_console_panel_choice_row}
\skillname{self_assembling_hover_opening_day_timeline}
\skillname{unfolding_button_into_springy_progress_track}
\skillname{layered_motion_build_up_walkthrough}
\skillname{pointer_swelling_edge_icon_strip}
\skillname{frosted_glass_now_playing_panel}
\end{skilllist} &
\resultpanel{p-33}{no_skill} &
\resultpanel{p-33}{general_skill} &
\resultpanel{p-33}{visual_skill} \\
\midrule

\instructionpanel{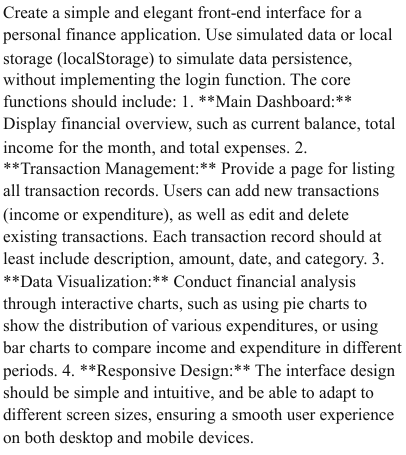} &
\begin{skilllist}
\skillname{water_shaped_period_row_with_in_column_time_scrub}
\skillname{glance_board_with_looping_edge_light}
\skillname{unit_mark_span_progress_with_forward_preview}
\skillname{synced_plotted_figure_index_console}
\skillname{gliding_focus_marker_with_self_drawing_confirmations}
\skillname{pointer_swelling_edge_icon_strip}
\end{skilllist} &
\resultpanel{p-35}{no_skill} &
\resultpanel{p-35}{general_skill} &
\resultpanel{p-35}{visual_skill} \\
\midrule

\instructionpanel{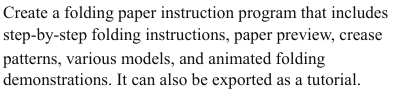} &
\begin{skilllist}
\skillname{continuous_fold_progress_model_viewer}
\skillname{rippling_sample_sheet_with_finish_swatches}
\skillname{layered_motion_build_up_walkthrough}
\skillname{cover_opens_into_full_window_page_turn_preview}
\end{skilllist} &
\resultpanel{p-45}{no_skill} &
\resultpanel{p-45}{general_skill} &
\resultpanel{p-45}{visual_skill} \\
\midrule

\instructionpanel{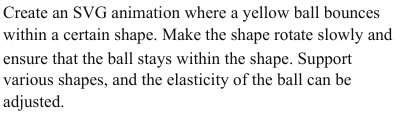} &
\begin{skilllist}
\skillname{tumbling_transparent_shell_with_bouncing_spheres}
\skillname{dot_swarm_shape_cycler}
\skillname{squash_and_bounce_mascot_loop_with_orbit_stage}
\skillname{fuzzy_shoveable_ball_in_bare_room}
\skillname{dormant_panel_that_wakes_and_layers}
\skillname{velocity_driven_glowing_smoke_trail}
\end{skilllist} &
\resultpanel{p-56}{no_skill} &
\resultpanel{p-56}{general_skill} &
\resultpanel{p-56}{visual_skill} \\
\midrule

\instructionpanel{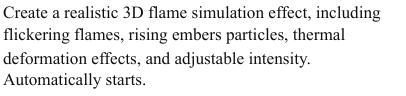} &
\begin{skilllist}
\skillname{drag_ignited_orbiting_flame_chamber}
\skillname{drag_seeded_rising_bubble_field}
\skillname{adjustable_open_water_scene}
\end{skilllist} &
\resultpanel{p-57}{no_skill} &
\resultpanel{p-57}{general_skill} &
\resultpanel{p-57}{visual_skill} \\
\midrule

\instructionpanel{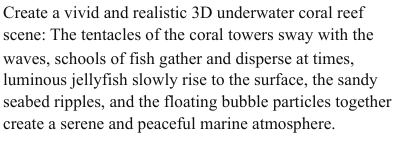} &
\begin{skilllist}
\skillname{submerged_drifting_creature_field}
\skillname{cursor_bent_drifting_particle_current}
\end{skilllist} &
\resultpanel{p-58}{no_skill} &
\resultpanel{p-58}{general_skill} &
\resultpanel{p-58}{visual_skill} \\
\midrule

\instructionpanel{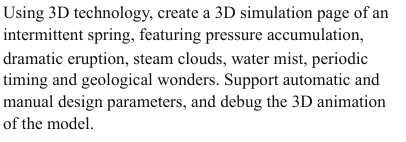} &
\begin{skilllist}
\skillname{orbitable_volumetric_plume_stage}
\skillname{adjustable_open_water_scene}
\skillname{orbitable_rippling_water_with_light_caustics}
\skillname{pointer_bound_square_cutaway_over_turning_subject}
\skillname{circling_point_with_linked_wave_traces}
\skillname{refracting_air_scene_with_terrain_buried_headline}
\skillname{drifting_morphing_cloud_sky}
\skillname{cursor_bent_drifting_particle_current}
\skillname{scroll_scrubbed_frame_sequence_stage}
\end{skilllist} &
\resultpanel{p-59}{no_skill} &
\resultpanel{p-59}{general_skill} &
\resultpanel{p-59}{visual_skill} \\
\midrule

\instructionpanel{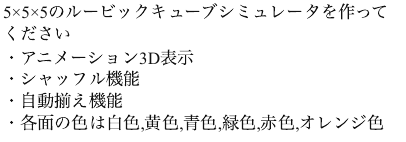} &
\begin{skilllist}
\skillname{orbitable_pressable_object_stage}
\skillname{articulated_subject_posing_stage}
\skillname{layered_motion_build_up_walkthrough}
\skillname{pressable_dimensional_keyboard_with_live_text_panel}
\end{skilllist} &
\resultpanel{p-62}{no_skill} &
\resultpanel{p-62}{general_skill} &
\resultpanel{p-62}{visual_skill} \\
\midrule

\instructionpanel{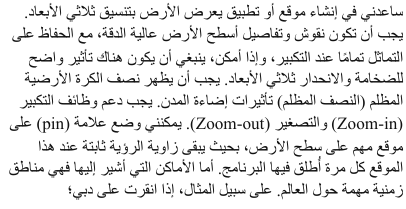} &
\begin{skilllist}
\skillname{illustrated_scene_nook_explorer}
\skillname{trailing_glass_magnifier_cursor}
\skillname{block_assembling_cursor_glossary_panel}
\end{skilllist} &
\resultpanel{p-63}{no_skill} &
\resultpanel{p-63}{general_skill} &
\resultpanel{p-63}{visual_skill} \\
\end{longtable}
\endgroup

\section{More Experimental Setup Details}
\label{app:setup_details}

\subsection{Visual Skill Library and Initial DAG}
\begin{table}[h]
    \centering
    \small
    \setlength{\tabcolsep}{3pt}
    \renewcommand{\arraystretch}{1.04}

    \caption{\textbf{Initial Skill--DAG statistics.}
    A Skill may belong to multiple Categories.}
    \label{tab:initial_skill_dag}

    \begin{tabularx}{0.72\linewidth}{@{}L{0.52\linewidth}
        >{\raggedleft\arraybackslash}X@{}}
        \toprule
        \textbf{Initial Category} & \textbf{Memberships} \\
        \midrule
        Dimensional Subject Showcase       & 78  \\
        Ambient Decorative Surface Layer   & 85  \\
        Presented Visual Media Viewing     & 256 \\
        Information Graphic Visualization & 21  \\
        Explorable Place and Environment   & 23  \\
        Expressive Interface Control Moment & 74 \\
        Visible Transition and State Change & 168 \\
        Self-Acting and Reactive Subject   & 140 \\
        Generative Visual Output Workshop  & 39  \\
        Hands-On Operable Live Field       & 56  \\
        First-Person View-Altering Overlay & 3   \\
        \midrule
        \textbf{Total memberships} & \textbf{943} \\
        \textbf{Unique Skills}     & \textbf{864} \\
        \bottomrule
    \end{tabularx}
\end{table}

We collect implementation-accessible examples from the
\href{https://tympanus.net/codrops/hub/all}{Codrops Creative Hub} and
\href{https://v0.app/templates}{v0 Templates}. After filtering and
deduplication, 850 source websites yield 864 Visual Skills, including 677 from
Codrops and 187 from v0 Templates. A source may yield multiple Skills when it
contains several independently reusable mechanisms. The Initial DAG contains
11 root Categories and 943 Category-to-Skill memberships; Table~\ref{tab:initial_skill_dag}
reports their distribution.

\mypara{Codrops sources.}
\href{https://tympanus.net/codrops/about/}{Codrops} is a web design and
development publication that hosts creative front-end demos. We use only
Codrops demos licensed under the
\href{https://tympanus.net/codrops/licensing/}{MIT License}.

\mypara{v0 sources.}
\href{https://v0.app/docs}{v0} is an AI-powered platform for building websites
and applications. We use only \href{https://v0.app/docs/templates}{public v0
templates}, which are published chats that other users can view, fork, and use.

\subsection{Potential Queries and Evolution Protocol}
Claude Opus 5 synthesizes 50 Skill-conditioned Potential Queries from each
initial Skill name and description, yielding 43,200 queries. We encode them
with Qwen3-Embedding-0.6B~\citep{zhang2025qwen3embedding} and fit a diagonal
Gaussian mixture model for each Skill, selecting the number of components by
the Bayesian information criterion over $K\in\{2,\ldots,6\}$. The nested
optimization sets contain 216, 432, 864, and 1,728 representative Queries,
corresponding to 0.25, 0.5, 1, and 2 Q/Skill. The Router, Reviewer, Diagnoser,
and Editor prompts are fixed across these checkpoints; only the accumulated
Potential-Query evidence and accepted graph or Skill edits change.

\subsection{SFT Training Configuration}
\label{app:sft_training_configuration}
Table~\ref{tab:sft_training_configuration} reports the supervised fine-tuning
settings.
\begin{table}[H]
    \centering
    \caption{\textbf{Key SFT training configuration for Qwen3.8-27B.}}
    \label{tab:sft_training_configuration}
    \footnotesize
    \setlength{\tabcolsep}{3pt}
    \renewcommand{\arraystretch}{1.12}
    \begin{tabularx}{\linewidth}{@{}L{0.35\linewidth}X@{}}
        \toprule
        \textbf{Parameter} & \textbf{Configuration} \\
        \midrule
        Base model & Qwen3.8-27B \\
        Numerical precision & bfloat16 training with FP32 master parameters \\
        Sequence length & 262,144 tokens \\
        Sequence packing & Enabled; packing length 262,144; padding-free batches \\
        Batch size & Micro batch size 1; global batch size 16 \\
        Learning rate & $2\times10^{-5}$; constant schedule; no warmup \\
        Regularization & Weight decay 0.1; gradient clipping 1.0 \\
        Adam parameters & $\beta_1=0.9$, $\beta_2=0.95$, $\epsilon=10^{-8}$ \\
        Loss scaling & \texttt{default+ignore\_empty\_think} \\
        Parallel topology & 2 nodes $\times$ 8 H800 GPUs; TP=8, PP=1, CP=2, DP=1 \\
        Random seed & 42 \\
        \bottomrule
    \end{tabularx}
\end{table}

\subsection{Benchmark Configurations}
\label{app:benchmarks}
Table~\ref{tab:web_generation_benchmark_protocols} summarizes the evaluated sets, browser interaction procedures, and scoring protocols.

\mypara{Interactive evaluation.}
All three benchmarks evaluate the generated websites through browser
interaction, but they assign the interaction and scoring roles differently.
For WebGen-Bench and Cookie-Bench, Seed2.0
Lite~\citep{bytedanceseed2026seed20} performs the browser interaction, while
GPT-5.4~\citep{openai2026gpt54} and GPT-5 serve as their respective scorers.
MiniAppBench instead uses a single GPT-5.4 agent for the complete evaluation
trajectory. The agent uses Playwright to open the generated website, inspect
its DOM, text, and CSS properties, perform clicks and form inputs, observe the
resulting page changes, and directly return the Intention, Static, and Dynamic
scores.

\mypara{Metric selection.}
We retain the benchmark-native measurements whose evidence matches the
interactive behavior evaluated in this work.
\begin{itemize}[leftmargin=3mm]
\item \underline{WebGen-Bench}~\citep{lu2025webgenbench}
reports functional Accuracy over 647 UI test cases and an additional
Appearance score from the initial rendered screenshot. We report Accuracy
and omit Appearance because a static view cannot observe visual mechanisms
that unfold through scrolling, hovering, state transitions, or temporal
animation.

\item
\underline{MiniAppBench}~\citep{zhang2026miniappbench}
defines Pass Rate using a weakest-dimension rule: an application passes only
when its Intention, Static, and Dynamic scores all reach $\tau=0.8$. This
threshold was calibrated for Gemini-3-Pro-Preview using a 200-item
expert-annotated subset, while our evaluation uses GPT-5.4. We therefore
report the three continuous dimension scores without transferring an
uncalibrated decision boundary.

\item
\underline{Cookie-Bench}~\citep{yang2026cookiebench}
first performs Static Perception and then applies Dynamic Scoring after
agent-driven interaction. We report the final Functionality, Aesthetics, and
Total scores from the \texttt{interactive\_video} protocol. These scores
incorporate the initial static evidence together with runtime logs,
screenshots, interaction traces, and continuous video, allowing the judge to
revise its provisional assessment after observing the website's behavior.
\end{itemize}
\begin{table*}[t]
    \centering
    \caption{\textbf{Configurations and evaluation protocols for the three website
    generation benchmarks.}}
    \label{tab:web_generation_benchmark_protocols}
    \footnotesize
    \setlength{\tabcolsep}{2pt}
    \renewcommand{\arraystretch}{1.3}

    \begin{tabularx}{\textwidth}{
        @{}
        >{\raggedright\arraybackslash}p{0.14\textwidth}
        *{3}{>{\raggedright\arraybackslash}X}
        @{}
    }
        \toprule
        \textbf{Aspect}
        & \textbf{WebGen-Bench}
        & \textbf{MiniAppBench}
        & \textbf{Cookie-Bench} \\
        \midrule

        Evaluated set
        & All 101 instructions, comprising 647 test cases
        & The 100-task validation split
        & The 844-task website subset \\

        Dynamic interaction
        & A WebVoyager agent executes each specified test operation and
        records the resulting page state
        & The scoring agent itself uses Playwright to inspect the page,
        perform clicks and form inputs, and observe page-state changes
        & A separate Cookie interaction agent autonomously explores the page
        through an observe--reason--act loop and records the interaction
        trajectory \\

        Interaction model
        & Seed2.0 Lite\newline
          \citep{bytedanceseed2026seed20}
        & GPT-5.4\newline
          \citep{openai2026gpt54}
        & Seed2.0 Lite\newline
          \citep{bytedanceseed2026seed20} \\

        Scoring model
        & GPT-5.4\newline
          \citep{openai2026gpt54}
        & GPT-5.4\newline
          \citep{openai2026gpt54}
        & GPT-5\newline
          \citep{openai2025gpt5} \\

        Scoring evidence
        & Original task instruction, test operation, expected outcome,
        action trajectory, and final screenshots
        & Original task instruction, Eval-Ref, generated code, DOM and CSS
        inspection, action results, and observed page-state changes
        & Original task instruction, source code, console logs, rendered
        screenshots, interaction video, action log, and static-stage scores \\

        Evaluation metrics
        & Accuracy,
        $\left(N_{\mathrm{yes}} + 0.5N_{\mathrm{partial}}\right) /
        N_{\mathrm{total}}$
        & Intention, Static, and Dynamic
        & Functionality, Aesthetics, and Total \\

        \bottomrule
    \end{tabularx}
\end{table*}

\subsection{Human Evaluation Protocol and Results}
\label{app:human_protocol}

We conduct a blinded pairwise preference evaluation on the website generation
results corresponding to the main experiment (Table~\ref{tab:web_generation_plugin_results}).
The evaluation compares matched Base and \textsc{+WebViSE} websites generated
from the same original instruction.

For each evaluation setting, we randomly sample 50 candidate pairs, resulting
in 350 pairwise comparisons across all evaluated settings. Four human
evaluators participated in the evaluation. Candidate pairs were distributed
among evaluators, and each pair received a single preference judgment from one
evaluator.

For each comparison, the evaluator sees only the original user instruction and
two anonymized websites, labeled Website A and Website B. Model identities,
generation conditions, and retrieved Visual Skills are hidden, and the A/B
order is randomized independently. The evaluator opens and interacts with both
websites before responding, inspecting their main content and available
scrolling, hovering, clicking, pointer interactions, state transitions, and
animations. The judgment considers alignment with the original instruction,
visual quality and coherence, and the usefulness and fidelity of the
interactions. Evaluators are instructed not to favor a website solely because
it contains more animation or visual effects.

After exploring both websites, the evaluator answers:
\emph{Considering the original user request, visual design quality, and
interactive experience, which website do you prefer overall?}

\begin{figure}[t]
    \centering
    \includegraphics[width=\linewidth]{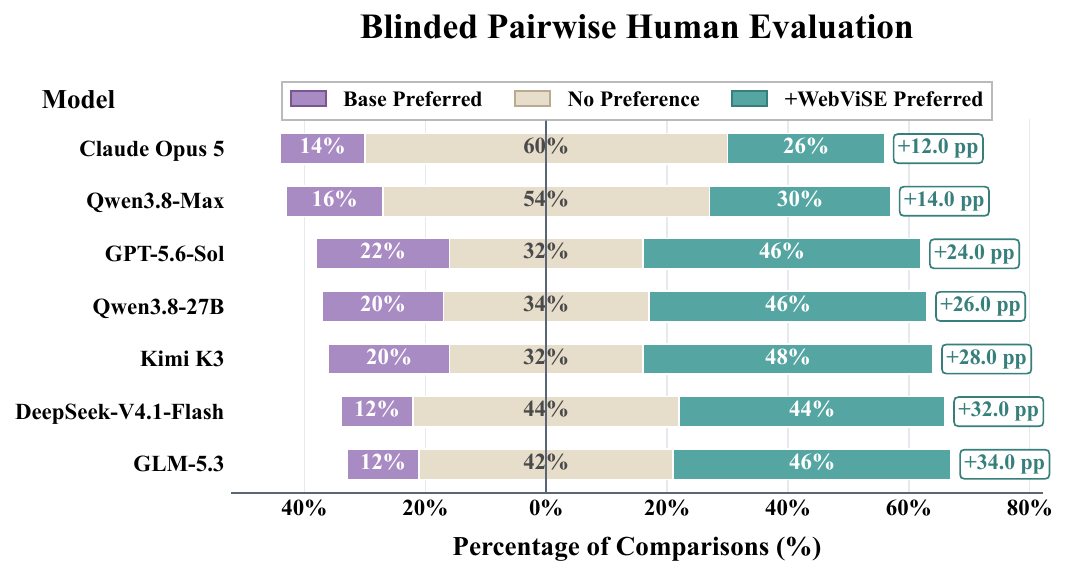}
    \caption{\textbf{Blinded pairwise human evaluation of Base and \textsc{+WebViSE}
    outputs across seven Web Generators.}}
    \label{fig:human_evaluation_pairwise_preference}
\end{figure}

The three choices are Website A, No Preference, and Website B. No Preference
covers comparable outputs, including cases where both are satisfactory or both
are poor. We restore the hidden condition labels after evaluation and report
the percentages of Base Preferred, No Preference, and \textsc{+WebViSE}
Preferred votes. Each percentage is computed from the corresponding pairwise
judgments. The preference margin in
Figure~\ref{fig:human_evaluation_pairwise_preference} is
$P_{\textsc{+WebViSE}}-P_{\mathrm{Base}}$ in percentage points.

Figure~\ref{fig:human_evaluation_pairwise_preference} shows that WebViSE is
preferred more often than Base for every evaluated generator, with preference
margins ranging from 12 to 34 percentage points. These results confirm that
WebViSE's improvements are perceptible to human evaluators.

\section{Additional Experimental Evaluation of WebViSE for Website Generation}
\label{app:additional_web_generation_evaluation}

\subsection{Detailed WebGen-Bench Functional Results}
\label{app:webgenbench_functional_results}
\begin{table}[h]
    \centering
    \caption{\textbf{Detailed WebGen-Bench functional results over 647 official UI
    test cases.} Yes, Partial, No, and Start Failed are case counts; Accuracy
    is reported as a percentage.}
    \label{tab:webgenbench_functional_breakdown}
    \footnotesize
    \setlength{\tabcolsep}{0.8pt}
    \renewcommand{\arraystretch}{1.0}
    \renewcommand{\tabularxcolumn}[1]{m{#1}}
    \resizebox{0.95\linewidth}{!}{%
    \begin{tabularx}{\linewidth}{@{}>{\raggedright\arraybackslash}m{0.21\linewidth}>{\raggedright\arraybackslash}m{0.115\linewidth}@{\hspace{2pt}}*{5}{Y}@{}}
        \toprule
        Model & Setting & Yes & Partial & No & \shortstack{Start\\Failed}
        & Accuracy (\%) \\
        \midrule
        \multirow{2}{=}{Claude Opus 5\\~\citep{anthropic2026opus5}}
        & Base & 245 & 245 & 157 & 0 & 56.80 \\
        & \textsc{+WebViSE} & 262 & 253 & 132 & 0 & 60.05 \\
        \midrule
        \multirow{2}{=}{Kimi K3\\~\citep{kimiteam2026kimik3}}
        & Base & 239 & 237 & 171 & 0 & 55.26 \\
        & \textsc{+WebViSE} & 219 & 279 & 149 & 0 & 55.41 \\
        \midrule
        \multirow{2}{=}{Qwen3.8-Max\\~\citep{qwen2026qwen38max}}
        & Base & 206 & 280 & 161 & 0 & 53.48 \\
        & \textsc{+WebViSE} & 220 & 281 & 146 & 0 & 55.72 \\
        \midrule
        \multirow{2}{=}{GLM-5.3\\~\citep{zhipuai2026glm53}}
        & Base & 196 & 260 & 191 & 0 & 50.39 \\
        & \textsc{+WebViSE} & 184 & 248 & 215 & 0 & 47.60 \\
        \midrule
        \multirow{2}{=}{DeepSeek-V4.1-Flash\\~\citep{deepseekai2026deepseekv41flash}}
        & Base & 233 & 286 & 128 & 0 & 58.11 \\
        & \textsc{+WebViSE} & 247 & 273 & 127 & 0 & 59.27 \\
        \midrule
        \multirow{2}{=}{GPT-5.6-Sol Standard\\~\citep{openai2026gpt56}}
        & Base & 178 & 243 & 226 & 0 & 46.29 \\
        & \textsc{+WebViSE} & 193 & 277 & 177 & 0 & 51.24 \\
        \midrule
        \multirow{2}{=}{Qwen3.8-27B\\~\citep{qwen2026qwen3827b}}
        & Base & 216 & 252 & 179 & 0 & 52.86 \\
        & \textsc{+WebViSE} & 221 & 262 & 164 & 0 & 54.40 \\
        \midrule
        Qwen3.8-27B NoSkill-SFT\\~\citep{qwen2026qwen3827b}
        & \raisebox{\baselineskip}[0pt][0pt]{Base}
        & \raisebox{\baselineskip}[0pt][0pt]{227}
        & \raisebox{\baselineskip}[0pt][0pt]{249}
        & \raisebox{\baselineskip}[0pt][0pt]{161}
        & \raisebox{\baselineskip}[0pt][0pt]{10}
        & \raisebox{\baselineskip}[0pt][0pt]{54.33} \\
        \midrule
        Qwen3.8-27B WebViSE-SFT\\~\citep{qwen2026qwen3827b}
        & \raisebox{\baselineskip}[0pt][0pt]{Base}
        & \raisebox{\baselineskip}[0pt][0pt]{218}
        & \raisebox{\baselineskip}[0pt][0pt]{257}
        & \raisebox{\baselineskip}[0pt][0pt]{172}
        & \raisebox{\baselineskip}[0pt][0pt]{0}
        & \raisebox{\baselineskip}[0pt][0pt]{53.55} \\
        \bottomrule
    \end{tabularx}%
    }
\end{table}

Table~\ref{tab:webgenbench_functional_breakdown} reports the complete
WebGen-Bench functional breakdown over all 647 official UI test cases. Following
the benchmark protocol, the Yes, Partial, No, and Start Failed counts partition
the test cases, while Accuracy is computed as
$\left(N_{\mathrm{yes}}+0.5N_{\mathrm{partial}}\right)/N_{\mathrm{total}}$.
The Base and \textsc{+WebViSE} settings follow
Table~\ref{tab:web_generation_plugin_results}, and \textsc{+WebViSE} websites
are still evaluated against the original task instructions. Here, Base denotes
inference from the original task instruction without retrieved Skills. The
two Qwen3.8-27B SFT variants are likewise evaluated from the original task
instructions without Skill retrieval.

\begin{figure}[t]
    \centering
    \includegraphics[width=0.88\linewidth]
    {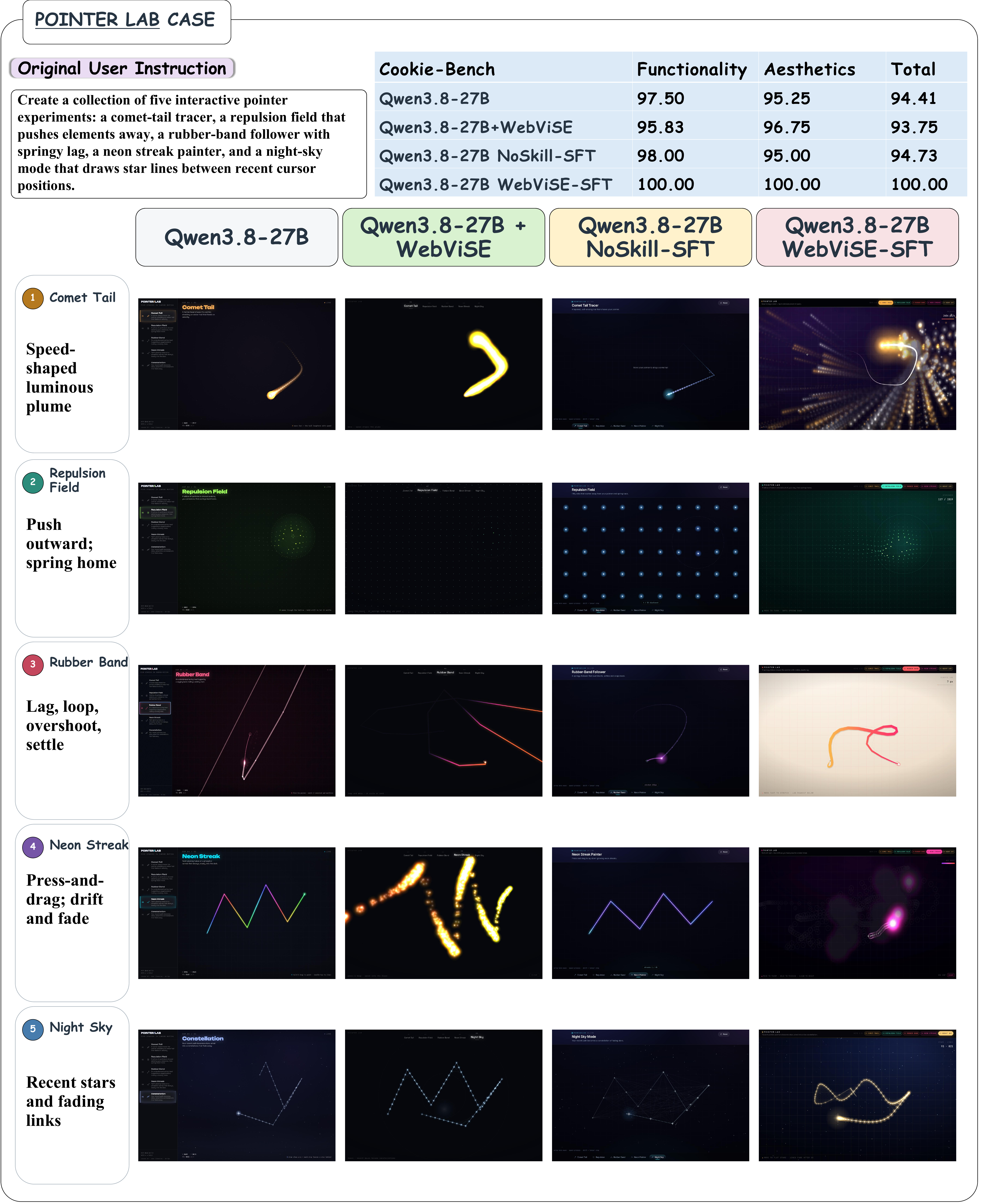}
    \caption{\textbf{Visual Skill internalization case study on Cookie-Bench.}
    Five pointer-responsive mechanisms from the same Original User Instruction
    are compared across Qwen3.8-27B, \textsc{+WebViSE}, NoSkill-SFT, and
    WebViSE-SFT. \textbf{For the full interactive experience, we encourage
    readers to explore the live
    \href{https://yifeisun-0710.github.io/WebViSE/internalization-case-study/base-original/}{Base},
    \href{https://yifeisun-0710.github.io/WebViSE/internalization-case-study/base-expanded/}{\textsc{+WebViSE}},
    \href{https://yifeisun-0710.github.io/WebViSE/internalization-case-study/no-skill-sft/}{NoSkill-SFT},
    and
    \href{https://yifeisun-0710.github.io/WebViSE/internalization-case-study/internalized/}{WebViSE-SFT}
    websites.}}
    \label{fig:pointer_lab_internalization_case_study}
\end{figure}

\subsection{Visual Skill Internalization Case Study}
\label{app:pointer_lab_internalization_case_study}

Figure~\ref{fig:pointer_lab_internalization_case_study} compares four settings
from the same Original User Instruction across five pointer mechanisms.
\textsc{+WebViSE} raises Aesthetics from 95.25 to 96.75, while WebViSE-SFT
scores 100 on all metrics and gains 5 Aesthetics points over NoSkill-SFT. Its
outputs pair each behavior with coherent styling: dimensional particles,
restrained spatial fields, contrasting trajectories, layered neon glow, and a
unified star-field palette. Since inference uses only the original
instructions, the result shows that aesthetic treatment and interaction
mechanisms are jointly internalized.

\section{Additional Skill Retrieval Evaluation}
\label{app:additional_evaluation}

\subsection{Retrieval Benchmarks and Protocol}
\label{app:retrieval_protocol}

\mypara{Evaluation benchmarks.}
We evaluate Skill retrieval on two frozen benchmarks.
\begin{itemize}[leftmargin=3mm]
\item The \textbf{Internal benchmark} contains 850 website-derived Goal
Queries with 864 source--Skill assignments and measures source--Skill
recovery. A Query may be associated with multiple source Skills, in which case
all associated Skills are treated as valid positives. Each Query retains the
source website's subject, purpose, and required functionality while withholding
the visual mechanism expressed by its source Skill. These Goal Queries also
inform the usage contexts of candidate Skills during library construction.

\item The \textbf{External benchmark} contains 119 English Goal Queries
independently collected from realistic website-building intents. These Queries
are neither derived from the source websites nor synthesized from Skill
documents, enabling evaluation of real-world generalization. Each pooled
Query--Skill pair receives a utility label in $\{0,1,2\}$, where 0 denotes
irrelevant, 1 partially useful, and 2 directly useful for realizing the Query.
The pairs are independently pre-annotated by Claude Opus
5~\citep{anthropic2026opus5}, Qwen3.8-Max~\citep{qwen2026qwen38max}, GPT-5.6
Sol~\citep{openai2026gpt56}, DeepSeek-V4.1-Flash~\citep{deepseekai2026deepseekv41flash},
GLM-5.3~\citep{zhipuai2026glm53}, and Kimi K3~\citep{kimiteam2026kimik3}. A
label is retained when at least four of the six judges agree on a unique rating.
Cases without consensus undergo anonymous human review, with human decisions
assigned greater weight in the final relevance judgments.
\end{itemize}
Both benchmarks are frozen before DAG evolution and used only for retrieval
evaluation thereafter. The Internal benchmark measures source--Skill recovery;
the External benchmark evaluates retrieval on intents independent of Skill
construction.

\mypara{Evaluation protocol and metrics.}
For each Query, a method returns a ranked list of Skills. An optimized node is
matched to its source Skills through retained canonical provenance, while each
static baseline node corresponds to one canonical Skill. Internal provides one
or more source Skills per Query, whereas External provides graded Skill-utility
labels. We therefore report MRR and Recall@$K$ for
$K\in\{1,5,10\}$ on Internal, and nDCG@$K$ for $K\in\{5,10\}$ and Utility
Recall@10 on External. For $N$ Queries, nDCG is
\begin{equation}
\begin{aligned}
\mathrm{nDCG}@K
&=\frac{1}{N}\sum_{i=1}^{N}
  \frac{\mathrm{DCG}_i@K}{\mathrm{IDCG}_i@K},\\
\mathrm{DCG}_i@K
&=\sum_{r=1}^{K}\frac{2^{y_{i,r}}-1}{\log_2(r+1)},
\end{aligned}
\label{eq:retrieval_ndcg}
\end{equation}
where $y_{i,r}\in\{0,1,2\}$ is the utility label at rank $r$, and
$\mathrm{IDCG}_i@K$ is the ideal DCG for Query $i$. Utility Recall@10 is
\begin{equation}
\begin{aligned}
\mathrm{UtilityRecall}@10
&=\frac{1}{N}\sum_{i=1}^{N}
  \frac{|\mathcal{R}_i\cap\widehat{\mathcal{Y}}_i^{10}|}{|\mathcal{R}_i|},\\
\mathcal{R}_i&=\{s\mid y_{i,s}\geq 1\},
\end{aligned}
\label{eq:retrieval_utility_recall}
\end{equation}
where $\widehat{\mathcal{Y}}_i^{10}$ is the top-10 retrieved Skill set.

For each of Claude Opus 5 and Qwen3.8-Max, we compare Flat LLM access to all
864 Skills, the Initial one-level DAG, and the Evolved DAG at 1 Q/Skill. Skill
documents, prompts, candidate budgets, top-$K$, and evaluation data are held
fixed within each model. In addition to retrieval quality, we report
query-count-weighted average input/output tokens and estimated cost. We use the
official prices current at evaluation time for Qwen3.8-Max\footnote{Qwen3.8-Max:
\$1.65 per million input tokens and \$4.951 per million output tokens.
\href{https://www.alibabacloud.com/help/en/model-studio/qwen3-8-max}{Official
pricing}, accessed September 12, 2026.} and Claude Opus
5.\footnote{Claude Opus 5: \$5 per million input tokens and \$25 per million
output tokens. \href{https://www.anthropic.com/news/claude-opus-5}{Official
pricing}, accessed September 19, 2026.} Costs are normalized separately for
each model, with Flat LLM set to 100\%.

\subsection{Additional Retrieval Baselines}
\label{app:retrieval_baselines}

We compare representative baselines spanning lexical and dense retrieval,
Potential-Query modeling, continual key evolution, retrieve-to-rerank
retrieval, and direct LLM-based Skill selection. BM25~\citep{robertson1994bm25}
retrieves Skills through lexical matching over their names and descriptions,
while Dense Retrieval ranks Skills by the cosine similarity between Query and
Skill embeddings. Hybrid-RRF~\citep{cormack2009rrf} combines the BM25 and Dense
rankings through reciprocal rank fusion. PQR~\citep{kang2025pqr} synthesizes
Potential Queries for each Skill and fits a compact distribution over their
embeddings, allowing an input Query to match the latent intents associated with
that Skill. ERM~\citep{hu2026erm} accumulates correctness-gated Query-expansion
signals into persistent Skill-key embeddings, allowing subsequent Queries to
benefit from earlier retrieval experience without rewriting the original Skill
documents. SkillRouter~\citep{zheng2026skillrouter} first retrieves candidate
Skills using a bi-encoder and then applies a pairwise relevance model to rerank
their full Skill documents. Flat LLM uses an LLM to select and rank Skills
directly from the flat Skill library without hierarchical routing.

For a controlled comparison, Dense Retrieval, Hybrid-RRF, PQR, and ERM share
Qwen3-Embedding-0.6B~\citep{zhang2025qwen3embedding} as their embedding
backbone, while SkillRouter uses its officially released retriever and
reranker. PQR and ERM use the same Potential-Query pool as WebViSE.
Table~\ref{tab:retrieval_additional_baselines} reports the complete results.

\begin{table}[h]
    \centering
    \caption{\textbf{Additional non-generative retrieval baselines.} These methods do
    not invoke an LLM during retrieval and therefore have no associated LLM
    token or cost fields.}
    \label{tab:retrieval_additional_baselines}
    \scriptsize
    \setlength{\tabcolsep}{2.4pt}
    \renewcommand{\arraystretch}{1.04}
    \resizebox{\linewidth}{!}{%
    \begin{tabular}{@{}L{0.3\linewidth}rrrrrrr@{}}
        \toprule
        \multirow{2}{=}{Method}
        & \multicolumn{4}{c}{Internal source recovery}
        & \multicolumn{3}{c}{External Skill utility} \\
        \cmidrule(lr){2-5}\cmidrule(lr){6-8}
        & R@1 & R@5 & R@10 & MRR@10
        & nDCG@5 & nDCG@10 & Utility R@10 \\
        \midrule
        BM25~\citep{robertson1994bm25}
        & 0.1232 & 0.2350 & 0.3021 & 0.1743 & 0.1250 & 0.1226 & 0.1004 \\
        Dense Retrieval
        & 0.2047 & 0.3626 & 0.4391 & 0.2723 & 0.2387 & 0.2225 & 0.1278 \\
        Hybrid-RRF~\citep{cormack2009rrf}
        & 0.1938 & 0.3297 & 0.4144 & 0.2572 & 0.1878 & 0.1794 & 0.1252 \\
        PQR~\citep{kang2025pqr}
        & 0.2959 & 0.5094 & \textbf{0.6153} & 0.3928
        & \textbf{0.3880} & \textbf{0.3662} & \textbf{0.2140} \\
        ERM~\citep{hu2026erm}
        & 0.2165 & 0.3691 & 0.4459 & 0.2937 & 0.2531 & 0.2374 & 0.1365 \\
        SkillRouter~\citep{zheng2026skillrouter}
        & \textbf{0.3785} & \textbf{0.5391} & 0.5803 & \textbf{0.4510}
        & 0.1589 & 0.1653 & 0.1129 \\
        \bottomrule
    \end{tabular}%
    }
\end{table}

\subsection{Evolution Ablations}
\label{app:retrieval_evolution_ablation}

Table~\ref{tab:retrieval_evolution_ablation} varies the Potential-Query budget
and isolates the effects of DAG and Skill edits using Qwen3.8-Max. Even 0.25
Q/Skill improves all seven quality metrics over the Initial DAG. The 1-Q/Skill
checkpoint gives the best Internal MRR@10 and External nDCG@10 among the tested
budgets, whereas 0.25 Q/Skill gives the highest Utility Recall@10 and 0.5
Q/Skill the lowest estimated cost. At 1 Q/Skill, DAG-only editing accounts for
most of the joint ranking improvement, Skill-only editing yields smaller gains,
and joint editing is strongest across all seven quality metrics.

\begin{table}[H]
    \centering
    \caption{\textbf{Complete evolution ablations with Qwen3.8-Max.} Average
    input/output tokens are query-count-weighted over the
    Internal and External benchmarks. Costs are normalized to Flat LLM.}
    \label{tab:retrieval_evolution_ablation}
    \scriptsize
    \setlength{\tabcolsep}{2.0pt}
    \renewcommand{\arraystretch}{1.04}
    \resizebox{\linewidth}{!}{%
    \begin{tabular}{@{}L{0.14\linewidth}rrrrrrrcr@{}}
        \toprule
        \multirow{2}{=}{Setting}
        & \multicolumn{4}{c}{Internal source recovery}
        & \multicolumn{3}{c}{External Skill utility}
        & \multirow{2}{*}{Avg. input / output}
        & \multirow{2}{*}{Relative cost} \\
        \cmidrule(lr){2-5}\cmidrule(lr){6-8}
        & R@1 & R@5 & R@10 & MRR@10
        & nDCG@5 & nDCG@10 & Utility R@10 & & \\
        \midrule
        \multicolumn{10}{@{}l}{\textit{Potential-Query budget}} \\
        Flat LLM
        & 0.3756 & 0.6015 & 0.6879 & 0.4753
        & 0.3330 & 0.3255 & 0.1863 & 67,023 / 749 & 100.00\% \\
        Initial DAG
        & 0.4129 & 0.6032 & 0.6785 & 0.4987
        & 0.4728 & 0.4419 & 0.2579 & 30,222 / 831 & 47.23\% \\
        0.25 Q/Skill
        & 0.4518 & 0.6447 & 0.7203 & 0.5397
        & 0.5336 & 0.5025 & \textbf{0.2968} & 25,830 / 1,164 & 42.33\% \\
        0.5 Q/Skill
        & 0.4600 & 0.6535 & 0.7221 & 0.5468
        & 0.5487 & 0.5037 & 0.2891 & 22,184 / 1,327 & \textbf{37.77\%} \\
        1 Q/Skill
        & \textbf{0.4806} & 0.6671 & 0.7306 & \textbf{0.5637}
        & 0.5483 & \textbf{0.5095} & 0.2895 & 22,472 / 1,535 & 39.09\% \\
        2 Q/Skill
        & 0.4682 & \textbf{0.6735} & \textbf{0.7315} & 0.5553
        & \textbf{0.5506} & 0.5048 & 0.2839 & 23,532 / 1,594 & 40.88\% \\
        \midrule
        \multicolumn{10}{@{}l}{\textit{DAG and Skill edits at 1 Q/Skill}} \\
        Initial DAG
        & 0.4129 & 0.6032 & 0.6785 & 0.4987
        & 0.4728 & 0.4419 & 0.2579 & 30,222 / 831 & 47.23\% \\
        DAG-edit only
        & 0.4676 & 0.6556 & 0.7262 & 0.5528
        & 0.5392 & 0.4984 & 0.2720 & 22,511 / 1,530 & 39.12\% \\
        Skill-edit only
        & 0.4241 & 0.6341 & 0.7162 & 0.5188
        & 0.4884 & 0.4543 & 0.2743 & 31,009 / 923 & 48.76\% \\
        Joint
        & \textbf{0.4806} & \textbf{0.6671} & \textbf{0.7306} & \textbf{0.5637}
        & \textbf{0.5483} & \textbf{0.5095} & \textbf{0.2895}
        & 22,472 / 1,535 & \textbf{39.09\%} \\
        \bottomrule
    \end{tabular}%
    }
\end{table}

\subsection{Structural and Qualitative Analysis}
\label{app:retrieval_analysis}

\begin{figure}[t]
    \centering
    \includegraphics[width=\linewidth]{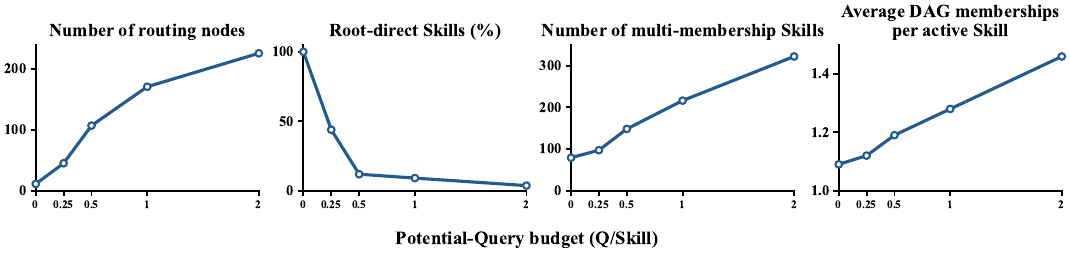}
    \caption{\textbf{Structural evolution of the Skill--DAG from the initial condition
    through four optimized Potential-Query budgets.} The four panels show non-redundant measures of graph
    growth, root-level attachment, cross-cutting multi-membership, and mean
    memberships per active Skill.}
    \label{fig:dag_structure}
\end{figure}

\mypara{Structural evolution.}
\label{app:structural_evolution}
At 1 Q/Skill, the graph grows from 11 to 171 routing nodes, root-direct Skills
decrease from 864 to 77, and multi-membership Skills increase from 79 to 216.
The mean memberships per active Skill rises from 1.09 to 1.28. The 864 source
Skill identities remain traceable through canonical provenance, and every
active Skill remains reachable. Table~\ref{tab:skill_dag_structure_stats} and
Figure~\ref{fig:dag_structure} provide the complete budget-wise statistics.

\begin{table}[h]
    \centering
    \caption{\textbf{Structural evolution across Potential-Query budgets.} Root-direct
    Skills are active Skills attached directly to an initial root Category.}
    \label{tab:skill_dag_structure_stats}
    \small
    \setlength{\tabcolsep}{3.5pt}
    \renewcommand{\arraystretch}{1.06}
    \resizebox{\linewidth}{!}{%
    \begin{tabular}{@{}L{0.16\linewidth}rrrrrr@{}}
        \toprule
        Budget & Routing nodes & Max depth & Root-direct Skills
        & Memberships / Skill & Multi-membership Skills & Active Skills \\
        \midrule
        Initial & 11 & 0 & 864 (100.0\%) & 1.09 & 79 & 864 \\
        0.25 Q/Skill & 45 & 1 & 378 (43.8\%) & 1.12 & 97 & 864 \\
        0.5 Q/Skill & 107 & 2 & 102 (11.8\%) & 1.19 & 148 & 864 \\
        1 Q/Skill & 171 & 2 & 77 (9.0\%) & 1.28 & 216 & 857 \\
        2 Q/Skill & 226 & 2 & 31 (3.6\%) & 1.46 & 322 & 857 \\
        \bottomrule
    \end{tabular}%
    }
\end{table}

\section{Complete Prompts}
\label{app:model_facing_prompts}
We provide the prompts used for Visual Skill construction,
query-induced Skill DAG evolution, and augmenting user instructions with
Visual Skills for plug-and-play website generation.
The input and output summaries describe the runtime payloads
and validated model responses in the released implementation.

\subsection{Website-Derived Visual Skill Construction}
The construction pipeline derives source-grounded Goal, Experience, and Interaction requests, distills Visual Skill candidates, produces source-confirmed \texttt{SKILL.md} artifacts, and serially constructs the Initial Skill DAG.

\subsubsection{Goal, experience, and interaction query construction}
\label{prompt:goal-experience-interaction-query-construction}
\begin{itemize}[leftmargin=3mm]
    \item \textbf{Input:} A source website implementation code.
    \item \textbf{Output:} Three incremental request segments describing the website's goal (G), intended experience (E), and interactions (I), saved in \texttt{query\_segments.json}.
\end{itemize}
\begin{promptbox}{Goal, experience, and interaction query construction}
Construct natural-language website requests from one provided website implementation and its observable behavior.

Working directory: {cwd}

Use only the supplied implementation and evidence. Identify the active behavior that establishes page purpose, content objects, user actions, maintained state, animation, transitions, rendering, and visible outcomes. Do not infer behavior from titles, tags, dependency lists, comments, or other metadata alone. If the supplied evidence does not establish a mechanism, omit it.

Do not describe the supplied evidence or extract a Web Skill. Use the evidence internally, then synthesize three incremental request segments: G, E, and I. Store them separately without duplicated earlier text so they can be concatenated as G, G+E, or G+E+I.

Except for G, a segment may contain multiple sentences when needed for adequate coverage. G is deliberately a terse, low-information user request. Every segment must add only the information assigned to its level and must not repeat, paraphrase, summarize, or contradict earlier segments.

Before writing, identify one dominant user task and its shortest complete experience/interaction path. Use that path as the backbone of all three segments. Omit secondary modes, alternate layouts/routes, autoplay, settings, optional ambience controls, and decorative micro-interactions unless they are indispensable to completing that dominant task or materially define the exhibit's central identity.

Critical separation rules:
- Write requests in natural user/product language, not source-analysis language.
- Do not copy the exhibit title or mention its author, brand, source identity, URLs, implementation symbols, CSS properties, shader variable names, equations, or source constants.
- G, E, and I must not name libraries, frameworks, APIs, rendering technologies, or other implementation labels.
- Do not name an inferred Skill, Skill, or taxonomy label.
- Do not prescribe low-level implementation.
- A request may describe an observable interaction or state change when appropriate.
- Preserve the page goal and core experience, not every decorative effect.
- Do not invent a business purpose, target audience, or domain that the source cannot support. When the source is only an effect/demo rather than a complete product page, formulate a generic request around the actual content object and demonstrated experience; do not turn it into an agency, portfolio, commerce, editorial, or campaign site without evidence.
- The concatenated queries must read naturally as standalone user requests. Avoid references such as "same as above", "as previously described", "this page", or "the source website".

Use the following coverage rules internally. They are selective checks, not output labels and not fields that must be populated without evidence. Include an item only when it is supported by the source and relevant to the dominant path.

1. G — goal:
   - Internally check: user, use context, core content objects, page goal, primary task, and intended outcome. These checks prevent misunderstanding; they are not a list to spell out in the output. State a user/domain only when indispensable and supported; otherwise remain generic rather than inventing one.
   - Use suitability for the user's tasks to decide what is central.
   - Output exactly one short, plain sentence or natural request fragment, normally about 5–25 English words. It should sound like the least-effort request a real person might type: directly name the wanted page/product and its single central purpose, then stop.
   - Prefer ordinary forms such as a bare noun phrase, "make/build/create/design ...", or a brief first-person request. Vary the surface form naturally across records. Do not make every request polished, explanatory, or complete like a product brief.
   - Do not enumerate all content fields, controls, menu destinations, secondary tasks, reasons, or expected benefits. Avoid explanatory padding such as "its only purpose is", "the aim is", "so visitors can", "coming away with", or contrasts describing what the page is not. Include a second clause only when the core task would otherwise be ambiguous.
   - Do not include or contrast layouts, spatial arrangements, visual effects, navigation behavior, movement, animation, or implementation details. Those belong only in later deltas.
   - Write G as a plain functional request from someone who has stated only the minimum desired artifact or task and has not chosen the experience design. Words such as immersive, interactive, full-screen, 3D, animated, spatial, floating, scrolling, dragging, hovering, looping, grid, or route normally indicate that information has leaked from a later level and must be removed.
   - G may draw only from the supported user/context, core objects, central task, and functional result, but it need not state all of them. A brief, broad quality word such as playful, calm, or atmospheric is allowed when it is a natural part of the user's minimal request and is supported by the source. Do not expand that word into a detailed experience description or stack several promotional adjectives; richer experience progression belongs in E.
   - A valid G for a gallery says that visitors should browse and appreciate a collection; it does not yet say that the collection floats in 3D, follows a route, or replaces a flat grid.

2. E — experience delta:
   - Check: dominant user path; changes in presented information and user attention; intended experience qualities; and the completion, exit, or return state when one exists.
   - Add stages such as entering, orienting, exploring, focusing, comparing, completing, or returning, but only when supported by the implementation.
   - Where evidenced, cover self-descriptiveness, conformity with user expectations, learnability, or user engagement.
   - Describe what the experience should feel like or make understandable, without describing input-response mechanics.
   - Do not mention device inputs, named controls, selectable modes, alternative layouts/routes, autoplay, sound toggles, settings, or other optional branches. Experience is the perceived progression of the dominant path, not a feature inventory.

3. I — interaction delta:
   - Check: primary input; affected object; state change; visible feedback; controllability; and reversal, interruption, or recovery when supported and relevant.
   - Express the complete causal chain from user action to visible response without naming the underlying algorithm or technology.
   - Where important, cover controllability and robustness to use error, including reversal, interruption, boundaries, or recovery.
   - Include only actions and responses forming the shortest complete dominant interaction path. Multiple related actions are allowed when the task inherently requires navigation, selection, inspection, and return, but omit autoplay, settings panels, layout/style/route variants, secondary controls, and decorative responses.

For G, write exactly one terse sentence or natural fragment, normally about 5–25 English words. For E and I, normally write two or three concise sentences and roughly 40–70 English words per segment; this is a soft target, not a reason to omit essential information.

Write exactly one UTF-8 JSON object to ./query_segments.json. Do not wrap it in Markdown. Use this schema:
{{
  "segments": [
    {{"level": "goal", "text": "..."}},
    {{"level": "experience", "text": "..."}},
    {{"level": "interaction", "text": "..."}}
  ]
}}

Validate that the file parses as JSON, has all three levels exactly once, and contains nonempty text. Write no other result files. After validation, reply with exactly DONE.

Before finishing, reopen and revise the file until all checks pass:
- G expresses only the minimum central artifact/task in one lazy, plain human utterance; the applicable Goal categories were checked internally rather than enumerated. A concise source-supported quality word is acceptable, but no detailed interaction, spatial-layout, motion, or mechanism information may leak into it.
- E adds only path, information/attention progression, experience qualities, and completion/return information; it contains no input-response description.
- I adds the primary input-to-visible-response chain and applicable control/reversal/recovery behavior, without optional controls.
- No segment repeats or paraphrases information already stated by an earlier segment.
- Concatenating the segments in order produces natural standalone requests.
\end{promptbox}

\subsubsection{Experience-category and Visual Skill candidate distillation}
\label{prompt:visual-skill-candidate-distillation}
\begin{itemize}[leftmargin=3mm]
    \item \textbf{Input:} One website's Goal, Experience, and Interaction request segments, without its source identity or code.
    \item \textbf{Output:} A JSON object containing Experience Category candidates, Visual Skill candidates, and their memberships; both candidate lists are empty when no eligible visual mechanism is found.
\end{itemize}
\begin{promptbox}{Experience-category and Visual Skill candidate distillation}
Extract query-friendly Experience Category candidates and concrete website Skill candidates.

The input included below contains one website's Goal, Experience and Interaction Queries. Source identity, source code and Skill library are intentionally unavailable.

Use all three Query levels with different responsibilities. Goal supplies the user's communication purpose, foreground content and plausible website-request contexts for Category retrieval and `Fits ...`. Experience supplies scope, composition, attention, progression and experiential qualities. Interaction supplies inputs, affected objects, state changes, visible feedback, boundaries and recovery. Goal context must not become a Category or Skill identity merely because it names a domain or page purpose. Abstract source-specific nouns after using them to understand legitimate applicability.

Two abstraction levels are required:

1. Experience Category answers: "What broad kind of impressive website effect is this useful for?" It is a coarse retrieval class a user with a short request could plausibly understand. Categories group by the primary content/use role, not by exact input, geometry, library or algorithm. A Category must be broad enough to contain many visibly different Skills across unrelated websites. Infer Category names from the supplied case rather than from a presumed taxonomy.

2. Skill answers: "What specific recognizable visual experience is offered inside that category?" A Skill is concrete and vivid enough to choose as a design direction, but it is not an implementation specification. Its complete document must preserve the visible composition, content roles, attention flow, meaningful experience progression when one exists, input-state-feedback chains, completion/return behavior, experience-defining constraints and portability boundary supported by Experience/Interaction. Infer its identity from the supplied case. Omit algorithms, libraries and ordinary components.

Visual-design eligibility is a hard gate. Emit a Skill only when the input states a concrete transferable visible treatment, transformation, spatial presentation, animation, reveal or interaction-to-visible-feedback relationship. Ordinary page structure, content organization, navigation, business flow, standard control behavior, loading or fallback behavior, implementation plumbing, and preservation of input, focus or other functional state are not visual Skills. An unnamed or unspecified visual-treatment slot is also insufficient: configurability and continuity may constrain a concrete effect but cannot replace the missing effect. Never invent an effect to ensure non-empty output.

If the input contains no eligible visual experience, use exactly this exclusion marker and emit no Category or Skill content:
{{
  "case_experience_summary": "NO_ELIGIBLE_VISUAL_SKILL",
  "category_candidates": [],
  "skill_candidates": []
}}

Use the smallest sufficient Category set. A cohesive experience normally needs one primary Category; add another only when the same complete Skill is genuinely useful under two distinct broad requests. Properties of mood, depth, duration, input, rendering, scale or inspection are normally Skill attributes rather than Categories. A foreground experience is not a background category unless it can actually function as a visual substrate behind foreground content. Domain and page-purpose labels are Goal types, not Experience Categories.

Do not use internal engineering, control-state or architecture abstractions as user-facing Category or Skill identities. Do not elevate ordinary interface support, readiness/status behavior or isolated implementation mechanisms into Skills. Keep coordinated content, visual layers, navigation, motion and feedback together when their relationship creates one recognizable experience. One cohesive experience should normally yield one complete Skill rather than fragments for each mechanism. Emit multiple Skills only for independently deployable experience regions or modes that remain recognizable when separated.

Remove brands, names, industries, source identity and incidental copy. Preserve content type only when it determines reuse. A single Skill may belong to multiple Categories. Do not pad Categories or split a complete experience into feature fragments. Every Category description must contain both a concise definition and a positive applicability sentence beginning with `Fits ...`. That sentence must name concrete, transferable website-request contexts in terms of communication purpose, foreground-content role or density, and placement in the experience. Do not satisfy it by merely repeating the Category's media object, visual material or mechanism. Do not add negative `Not for` wording. Keep Skill name and description concise for retrieval, but make the remaining Skill fields complete enough to guide implementation after selection. Do not replace missing evidence with generic web-design advice.

Each Category candidate must contain exactly two fields: `name` and `description`. Put every useful Category-level definition, applicability qualification, retrieval cue, clarification or other explanation into `description`; never create an additional field. Category–Skill membership is represented only from the Skill side through `skill_candidates[*].category_candidate_names`. Every emitted Category name must occur in at least one such list.

Use exactly:
{{
  "case_experience_summary": "short user-facing summary",
  "category_candidates": [
    {{"name": "lowercase_snake_case", "description": "broad use-oriented definition. Fits concrete transferable website-request contexts described by communication purpose, foreground content and placement."}}
  ],
  "skill_candidates": [
    {{
      "name": "lowercase_snake_case",
      "description": "specific recognizable visual experience in user-facing language",
      "experience_scope": "where this experience begins and ends and whether it is a region or whole-view experience",
      "content_and_composition": ["content roles, cardinality, spatial/layer relationships and persistent elements"],
      "experience_progression": ["meaningful ordered stages, continuous advance, looping or return; empty when the experience has no real progression"],
      "attention_and_information_flow": ["how prominence and revealed information change"],
      "interaction_model": [
        {{
          "input": "user input or autonomous driver",
          "affected_object": "content or visual role being changed",
          "state_change": "observable state transition",
          "visible_feedback": "what the user sees",
          "control_and_recovery": "reversal, interruption, resumption, recovery or boundary behavior"
        }}
      ],
      "experience_qualities": ["observable experiential qualities"],
      "completion_and_return": ["completion, looping, reset, return or persistent-state behavior"],
      "locked_observable_requirements": ["requirements that define this Skill and must survive transfer"],
      "adaptable_details": ["details that may change without changing Skill identity"],
      "portability_boundary": "what kinds of content/context fit and what change would create a different Skill",
      "category_candidate_names": ["category_name"]
    }}
  ]
}}

When candidates are emitted, every emitted Category name must be referenced by at least one Skill through `skill_candidates[*].category_candidate_names`, and every Skill must reference at least one emitted Category. The exclusion marker must be used only with both candidate lists empty. Return only the JSON object, without Markdown fences or commentary.
\end{promptbox}

\subsubsection{Website-derived Visual Skill construction}
\label{prompt:website-derived-visual-skill-construction}
\begin{itemize}[leftmargin=3mm]
    \item \textbf{Input:} A selected Visual Skill candidate and the corresponding website implementation source.
    \item \textbf{Output:} One or more reusable \texttt{SKILL.md} documents and a separate \texttt{skill\_manifest.json} recording their source-code evidence.
\end{itemize}
\begin{promptbox}{Website-derived Visual Skill construction}
Create reusable Web Skills for one already-selected website Skill by inspecting its implementation source.

Working directory: {cwd}

Use the supplied selected Skill as the experience boundary: extract only mechanisms that materially implement it. Inspect the supplied website source project directly. Treat metadata and dependency names only as navigation hints, never as evidence that a mechanism is active.

Write all final artifacts under ./output.

## Objective

Turn source-confirmed implementation knowledge into one or more portable Skills applicable to an unrelated website Query. Preserve the causal mechanisms needed to realize the Skill while removing the original subject, brand, assets, copy, routes, repository structure and application shell.

Skill granularity is not fixed. Prefer one coherent Skill when mechanisms share authoritative state, coordinate systems, timing, update order, rendering output or a transition lifecycle. Split only when each resulting mechanism is independently usable, independently triggered and still meaningful after the others are removed. Do not atomize one coordinated experience into small effects merely because its implementation has multiple modules. Do not bundle unrelated decoration that does not materially implement the selected Skill.

Do not promote generic collection routing, back navigation, loading and unsupported-environment notices, error handling, cleanup, ordinary controls or application-shell plumbing into standalone Skills. They may appear only as constraints inside a defining visual mechanism when required for that mechanism to remain usable. A fallback or ordinary interface behavior is a Skill only when the Skill's name and description make that behavior itself the distinctive designed experience; its presence in secondary requirement fields is insufficient.

Do not force compatibility with an existing Skill library. Generate the representation warranted by this Skill and source.

## Source analysis

Recursively inspect the active source. Trace entry points and relevant imports. Establish, from active code rather than filenames or comments:

- the mechanism goal and affected visual/content objects;
- inputs, autonomous drivers and initialization state;
- authoritative state and derived state;
- update, transition, simulation or rendering rules;
- coordinate conversions and synchronization boundaries;
- meaningful update order and lifecycle ownership;
- visible outputs and feedback;
- responsive behavior, performance constraints and cleanup;
- constants, equations, thresholds and timing relationships that materially define behavior.

Runtime inspection is optional. Screenshots are not required. Do not claim behavior that source relationships cannot establish.

## Generalization boundary

Write each Skill for future unrelated content. The Skill must explain how to implement the transferable mechanism, not how to reproduce this website.

- Generalize subject matter, copy, branding, route names, asset filenames, exact item content and incidental styling.
- Preserve a content type only when the mechanism structurally depends on it.
- Preserve specialized runtime libraries or low-level systems only when their active role materially determines the mechanism. Explain the role; never emit a dependency list.
- Omit application frameworks, component organization, rendering-delivery modes, build tools and source module structure. Do not prescribe React, Vue, Svelte, Angular, Astro, native-DOM architecture, JSX/TSX, SSR/CSR/SSG, Vite, Webpack or Parcel.
- Retain implementation-critical systems such as Three.js, GSAP, Lenis, WebGL/WebGPU, GLSL/WGSL, GPU instancing, shaders, physics, raycasting, animation mixers, video textures or post-processing only when verified and causally useful.
- Prefer parameter relationships and observable invariants over blindly copying source constants. Preserve exact values only when changing them would materially alter identity, correctness or synchronization.
- Do not mention the source, exhibit, repository, article, author, data ID, Skill ID or inspection process inside SKILL.md.

## SKILL.md contract

For every Skill, write `./output/skills/<name>/SKILL.md`.

The YAML frontmatter must contain exactly:

---
name: <lowercase kebab-case, at most 64 characters>
description: <one concise sentence stating when to use the Skill and the distinctive observable result>
---

Use imperative instructions. Keep the body focused on non-obvious implementation knowledge and below 500 lines. Use exactly these top-level sections after one H1 title:

- `## Use this skill when`
- `## Core mechanism`
- `## Implementation workflow`
- `## State and update model`
- `## Constraints and invariants`
- `## Adaptation points`
- `## Validation`

The body must be sufficiently operational without access to the source. Include equations, pseudocode, state transitions, coordinate relationships and update order only when they improve faithful implementation. State visible validation criteria, common causal failure modes and what may vary without changing the Skill.

Do not include provenance, source paths, original names, evaluation commentary, generic web advice, installation steps, README material or a reconstruction narrative.

## Manifest and evidence

Write `./output/skill_manifest.json` using exactly:

{{
  "decomposition_rationale": "why these Skills are the smallest coherent transferable units",
  "skills": [
    {{
      "name": "same frontmatter name",
      "description": "same frontmatter description",
      "relative_path": "skills/<name>/SKILL.md",
      "skill_contribution": "which defining part of the supplied Skill candidate this generated Skill implements",
      "mechanism_scope": ["coherent responsibilities kept together in this Skill"],
      "source_evidence": [
        {{"path": "relative source path", "evidence": "active code relationship establishing the mechanism"}}
      ]
    }}
  ]
}}

The manifest is provenance and may cite source paths. SKILL.md must remain source-neutral. Emit at least one Skill because the supplied Skill candidate was selected as an implemented experience, but do not target any particular Skill count.

Before finishing, validate every path, frontmatter pair and required section; confirm each Skill is source-grounded, jointly covers the selected Skill candidate's defining mechanism, and remains useful for unrelated website content. Reply with exactly DONE after the files are valid.
\end{promptbox}

\subsubsection{Serial Initial Skill DAG construction}
\label{prompt:serial-initial-skill-dag-construction}
\begin{itemize}[leftmargin=3mm]
    \item \textbf{Input:} One complete incoming Visual Skill and the complete active root Category index produced by all preceding Skills.
    \item \textbf{Output:} Existing Category selections, optional description revisions that preserve Category identity, and at most one new Category for an unmatched salient role. The program applies these decisions as Category-to-Skill edges before processing the next Skill.
\end{itemize}
\begin{promptbox}{Serial Initial Skill DAG construction}
Integrate one incoming Visual Skill into the current root Category index and update the one-level Initial Skill DAG.

The input contains one complete Visual Skill and the complete active Category index built from earlier Skills. Read the complete Skill, identify its independently central visual roles, and route it by broad query-facing purpose. A Category is a stable first-hop routing concept shared by many visibly different Skills; it is not an exact mechanism, implementation technology, source, page-industry label, mood, layout, input device, or isolated effect property.

Assign the Skill to every existing Category whose description genuinely covers an independently central role. Multiple memberships are valid when removing any selected role would materially change the kinds of user Queries the Skill can serve. Do not select multiple Categories for incidental objects, phases, or different properties of one coherent role.

Prefer existing Categories over index growth. If an existing Category has the correct identity but its current description is accidentally narrow, revise only its description into a broader positive query-facing definition without changing its name or identity. Do not broaden a Category across distinct affected-content families or visible functions, and do not merge, split, or rename existing Categories during Initial DAG construction.

If the most salient central role is not covered by any existing Category, add at most one new Category for that unmatched role. The new Category must express a broad affected-content family and visible function that will remain useful for routing substantially different future Skills. A new mechanism, medium, geometry, interaction, animation, subject, or visual style alone does not justify a new Category. When the index is empty, create the first Category from the incoming Skill's most salient broad visual role.

Every added or revised Category description must contain a concise definition followed by a positive applicability sentence beginning with `Fits ...`. Remove source identity, provenance, brands, implementation details, exact assets, and incidental subject matter. Keep concrete implementation knowledge in the Skill rather than copying it into Category identity.

Return exactly one JSON object:
{{
  "routes": [
    {{
      "operation": "select|revise",
      "category_name": "exact existing Category name",
      "revised_category": null
    }}
  ],
  "new_category": null,
  "rationale": "brief explanation of the central roles, matched Categories, and any unmatched salient role"
}}

For `select`, preserve the existing Category and set `revised_category` to null.
For `revise`, preserve `category_name` and provide:
{{"name":"the exact same existing Category name","description":"broadened definition. Fits transferable website-request contexts."}}

When one unmatched salient role requires a new Category, replace `new_category` with:
{{"name":"lowercase_snake_case","description":"broad definition. Fits transferable website-request contexts."}}

`routes` may be empty when no existing Category matches, but `new_category` must then create the Skill's membership. When the index is empty, `new_category` must create the first Category. Every result must assign the Skill through at least one existing or new Category. Use only supplied existing Category names, do not repeat a Category, and add no more than one new Category. Return only the JSON object without Markdown fences or commentary.
\end{promptbox}

\subsection{Query-Induced Joint Skill--DAG Evolution}
Query-induced Skill DAG evolution generates Skill-conditioned Potential Queries, retrieves Skills through shared DAG routing and applicability review, diagnoses retrieval failures, and performs structural partitioning, partition completion, and non-structural Skill editing. The diagnosis and editing prompts operationalize the four failure types described in the method: \texttt{alternative\_same\_position} evidence identifies mixed mechanism families and triggers a split that replaces broad parent memberships with narrower child memberships; \texttt{rewrite\_node} and \texttt{add\_edge} repair routing or access mismatches; \texttt{rewrite\_skill} repairs unclear Skill applicability; and \texttt{same} evidence identifies mechanism-equivalent merge candidates. Prompt~\ref{prompt:sparse-skill-dag-revision} first proposes structural partitions and is reused after partition completion for non-structural Skill edits. Merge candidates from the non-structural pass are verified from the complete Skill documents by Prompt~\ref{prompt:canonical-skill-merge-verification}.

\subsubsection{Skill-conditioned Potential Query generation}
\label{prompt:skill-conditioned-potential-query-generation}
\begin{itemize}[leftmargin=3mm]
    \item \textbf{Input:} The exact name and retrieval description of one Visual Skill.
    \item \textbf{Output:} A JSON array containing exactly 50 unique Goal-level website requests used as Potential Queries.
\end{itemize}
\begin{promptbox}{Skill-conditioned Potential Query generation}
Generate exactly 50 diverse Goal-level English Potential Queries for one visual Web Skill.

These queries will be used only to build an offline retrieval index. Infer plausible user goals from the supplied Skill name and description. Do not consult any held-out query, retrieval result, relevance label, or external library.

Requirements:
- Return exactly one JSON object: {{"potential_queries":["...", "...", ...]}}.
- The array must contain exactly 50 unique non-empty English strings.
- Each string must sound like a natural request from someone asking for a website or web experience, normally 4-35 words and never more than 50 words.
- Express the desired outcome, experience, content presentation, or user task. Do not explain the Skill and do not mention that a Skill is being retrieved.
- Do not copy the Skill name verbatim. Avoid library/framework names, implementation instructions, source filenames, numeric parameters, and internal mechanism terminology unless the user-facing concept is indispensable.
- Keep each query focused on one primary goal; avoid feature inventories and mini-PRDs.
- Make the set genuinely diverse, not 50 superficial synonym substitutions. Cover a balanced mixture of terse requests, outcome-oriented requests, audience/context variants, experience-oriented requests, and plausible page/product contexts.
- Do not number the strings and do not include Markdown or commentary.

Skill name: {name}
Skill description: {description}
\end{promptbox}

\subsubsection{DAG routing}
\label{prompt:dag-routing}
\begin{itemize}[leftmargin=3mm]
    \item \textbf{Input:} One website Query, the current routing depth, and the exact names, descriptions, and parent names of all candidate routing nodes at that depth.
    \item \textbf{Output:} A possibly empty ranked JSON list of every selected node, with a Query-grounded reason and confidence for each selection.
\end{itemize}
\begin{promptbox}{DAG routing}
Route one website Query through the supplied candidates at one level of a Skill DAG.

Select every node whose query-facing definition can materially support the requested visible experience, interaction, content presentation, or task. Preserve multiple plausible paths when the Query leaves the concrete visual mechanism open. Exclude nodes supported only by topic, mood, implementation stack, generic page type, or as insurance. Do not impose a selection quota. An empty ranking is valid when no candidate applies.

Use exactly:
{"ranking":[{"name":"exact supplied node name","reason":"brief Query-grounded reason","confidence":"high|medium|low"}]}

Use supplied names exactly and at most once. Rank selected nodes by applicability and return only the JSON object.
\end{promptbox}

\subsubsection{Full-document Skill applicability review}
\label{prompt:skill-applicability-review}
\begin{itemize}[leftmargin=3mm]
    \item \textbf{Input:} The original website Query and a batch of complete Visual Skill documents reached through DAG routing.
    \item \textbf{Output:} One applicability decision, integer score, and Query-grounded reason for every supplied Skill; the program ranks applicable Skills and retains the top-$K$ results.
\end{itemize}
\begin{promptbox}{Full-document Skill applicability review}
Review every supplied complete Visual Skill document for one website Query.

Judge the reusable visible mechanism in the complete document against the Query's requested content, object, state, task, visual outcome, and interaction outcome. A Skill is applicable when at least one non-trivial, recognizable mechanism can operate on the Query's existing subject or task without inventing unrelated business meaning. Ignore incidental source branding, copy, assets, frameworks, and surrounding workflow. Reject shared topic, mood, page type, routine polish, or mere technical insertability.

Decide every supplied Skill exactly once. Give an integer applicability score from 0 to 100; scores compare direct usefulness across candidates, not visual quality in isolation.

Use exactly:
{"decisions":[{"name":"exact supplied Skill name","applicable":true,"score":0,"reason":"brief Query-grounded applicability reason"}]}

Preserve supplied names and return only the JSON object.
\end{promptbox}

\subsubsection{Retrieval-failure diagnosis}
\label{prompt:retrieval-failure-diagnosis}
\begin{itemize}[leftmargin=3mm]
    \item \textbf{Input:} One representative Potential Query; its reference Skill; the observed candidate, rank, root, and descendant-node outcomes; the current routing-node index; and a co-retrieved Skill batch with memberships and observed ranks.
    \item \textbf{Output:} A JSON diagnosis containing one retrieval-miss attribution and any strongly supported mechanism-equivalence or alternative-same-position relations.
\end{itemize}
\begin{promptbox}{Retrieval-failure diagnosis}
Diagnose one observed retrieval trajectory from a current Skill-DAG checkpoint.

The source Skill is an observed positive because the Goal Query was generated from that Skill,
but source identity is not proof that it is equivalent to another Skill. Make only sparse,
future-query-valid edits.

Miss attribution:
- none: source already ranks within the supplied top-K retention boundary and no retrieval repair is needed.
- rewrite_skill: the source mechanism genuinely supports the Query, but its current short
  retrieval description lacks the ordinary query-facing bridge. Preserve its mechanism.
- rewrite_node: the source is already attached to the right node, but that node wording caused
  the router to miss it. Keep the node broad and coherent.
- add_edge: the source independently belongs to an existing node but lacks that membership.
  Never add an edge merely to rescue this one Query.

Respect the observed pipeline boundary. If the source is absent from the routed candidate set,
the failure is `root_route` and only rewrite_node/add_edge/none can repair it because Skill text
was never read. If the source is in the candidate set but outside the supplied top-K retention
boundary, the failure is `candidate_rank` and only rewrite_skill/none can repair it. A retained
source requires none/none.

Within the supplied co-retrieved batch emit only strong relations:
- same: the two Skills express the same reusable visible mechanism; adaptable details are the
  only differences. This can support merge after full-document verification.
- alternative_same_position: materially different mechanisms that can independently occupy the
  same Query-grounded visual/function position. Keep the Skills distinct. When they coexist
  under an over-broad parent, provide that parent and a concise routing distinction; repeated
  evidence can trigger split = remove old parent-to-Skill edges + add narrower edges.
- omit weak/topic/style/coexistence relationships.

Return exactly JSON:
{
  "miss_attribution": {
    "stage": "none|root_route|candidate_rank",
    "operation": "none|rewrite_skill|rewrite_node|add_edge",
    "target_name": "exact supplied Skill or node name, or empty string",
    "replacement_description": "complete replacement only for rewrite operations, else empty string",
    "reason": "brief evidence-grounded reason"
  },
  "relations": [
    {
      "a": "exact supplied Skill name",
      "b": "exact supplied Skill name",
      "relation": "same|alternative_same_position",
      "reason": "mechanism-level reason",
      "split_parent": "exact supplied common parent for alternative relation, else empty string",
      "routing_distinction": "query-facing distinction for alternative relation, else empty string"
    }
  ]
}
Use exact supplied names. JSON only.
\end{promptbox}

\subsubsection{Sparse joint Skill--DAG revision}
\label{prompt:sparse-skill-dag-revision}
\begin{itemize}[leftmargin=3mm]
    \item \textbf{Input:} One current parent node, its evidence-linked Skills, repeated miss and relation evidence aggregated across representative Potential Queries, and the program-computed split constraints for that parent.
    \item \textbf{Output:} A sparse JSON proposal containing any evidence-supported node rewrite, Skill rewrite, membership addition, merge candidate, or node split. The program validates and applies accepted operations separately.
\end{itemize}
\begin{promptbox}{Sparse joint Skill--DAG revision}
Produce a sparse query-induced revision for one current Skill-DAG parent node.
The input contains repeated evidence collected only from representative Potential Queries after running the
current checkpoint. Do not invent evidence and do not edit merely to rescue one Query.

Allowed decisions:
- Rewrite a Skill retrieval description only when repeated miss evidence shows that its concrete
  mechanism supports ordinary Goal wording that the current description fails to expose.
- Rewrite the parent description only when repeated root-route misses expose one coherent missing
  fit cue. Do not turn it into a catch-all.
- Add a parent-to-Skill membership only when repeated evidence shows the complete Skill
  independently belongs under this parent for future Queries. Concretely, an `add_edges` item is
  allowed only when `repeated_miss_evidence` contains `operation="add_edge"` for that exact Skill
  and current parent. A `same` relation alone never authorizes a new membership.
- Merge only a `same` pair supported by repeated independent Queries from both source directions.
  Same page type, mood, medium or rough purpose is insufficient.
- Split when repeated `alternative_same_position` evidence shows the parent mixes stable,
  query-distinguishable mechanisms. The alternative relation is evidence that the CURRENT PARENT
  is too broad; it is not a pairwise cannot-link constraint. Repartition the whole supplied set into
  small reusable mechanism families. Skills belong in the same family when they share the same
  causal kernel: trigger, controlled state, affected object, visible feedback, temporal progression,
  interruption/recovery rule, and portability boundary. Differences in item count, copy, imagery,
  page scale, optional flourish, or list-versus-single-item presentation do not by themselves justify
  separate families. Use 2-8 coherent children; never create one child per Skill.
- For a Skill already directly attached to this parent, a split removes that broad membership and
  adds the narrower child membership. A supplied evidence-linked Skill currently attached elsewhere
  may also enter a child family; keep its other valid memberships because the DAG is multi-parent.

The input includes `split_required`, `minimum_split_move_count`, and
`split_candidate_skill_names`. When `split_required` is true, return exactly one split, move at
least `minimum_split_move_count` listed candidates, and partition them into 2-8 coherent children.
When `splits_allowed` is false, return an empty `splits` list. This means the structural partition
has already been staged from the same diagnosis evidence and this call must only complete the
non-structural edits; it does not mean that the evidence is from a later evolution round.
This requirement is activated only after dense repeated alternative evidence; it is not permission
to invent a taxonomy. Use only the supplied Skill names/descriptions and the query-induced relation
signals. Do not infer families from source site, provenance, code, author, or implementation stack.
Before emitting children, compare all candidates jointly. Repeated `same` links should normally land
in one child even when a conservative full-document merge would retain separate Skill identities.

Prefer no edit over weak edit. Return exactly JSON:
{
  "parent_rewrite": null or {
    "replacement_description": "complete replacement",
    "evidence_query_ids": ["supplied id"]
  },
  "skill_rewrites": [{
    "skill_name": "supplied Skill",
    "replacement_description": "complete replacement",
    "evidence_query_ids": ["supplied id"]
  }],
  "add_edges": [{
    "skill_name": "supplied Skill",
    "target_node": "the current parent name",
    "evidence_query_ids": ["supplied id"]
  }],
  "merges": [{
    "keep_skill": "supplied Skill",
    "remove_skill": "supplied Skill",
    "evidence_query_ids": ["supplied id"]
  }],
  "splits": [{
    "parent_node": "the current parent name",
    "children": [{
      "name": "new unique snake_case node name",
      "description": "broad query-facing definition with a Fits sentence",
      "move_skill_names": ["supplied direct member Skill"]
    }],
    "evidence_query_ids": ["supplied id"]
  }]
}
JSON only.
\end{promptbox}

\subsubsection{Canonical Skill merge verification}
\label{prompt:canonical-skill-merge-verification}
\begin{itemize}[leftmargin=3mm]
    \item \textbf{Input:} Two complete canonical Visual Skill documents and the representative Potential-Query evidence identifiers supporting the proposed merge.
    \item \textbf{Output:} A JSON equivalence decision, the exact canonical Skill identity to retain when equivalent, and a brief mechanism-level reason.
\end{itemize}
\begin{promptbox}{Canonical Skill merge verification}
Verify whether two complete canonical Skills express the same reusable
visible mechanism. `same` requires mechanism-level equivalence: merging must not remove a material
behavior, progression, interaction, recovery rule, or portability boundary. Shared page type,
medium, style, mood, purpose, or being alternatives for one Query is not equivalence. Prefer false
when uncertain. If equivalent, choose the clearer/more complete canonical Skill to keep.
Return exactly JSON: {"equivalent":true|false,"keep_skill":"exact supplied name or empty string","reason":"brief mechanism-level reason"}. JSON only.
\end{promptbox}

\subsubsection{Partition assignment after node splitting}
\label{prompt:partition-assignment}
\begin{itemize}[leftmargin=3mm]
    \item \textbf{Input:} One split parent, the induced child-node definitions, and a batch of complete active Skill documents still attached directly to the parent.
    \item \textbf{Output:} A JSON assignment for every supplied Skill, selecting one or more exact child names or explicitly retaining the Skill at the parent.
\end{itemize}
\begin{promptbox}{Partition assignment after node splitting}
Complete one learned Skill-DAG routing partition.
The child nodes were induced from repeated representative Potential-Query evidence. For every supplied
remaining Skill, select the child node or nodes whose retrieval boundary materially covers the
Skill's complete reusable visible mechanism.

Rules:
- This is mechanism-family routing, not topical or page-form tagging. Match the causal kernel:
  trigger, controlled state, affected object, visible feedback, temporal progression,
  interruption/recovery, and portability boundary.
- Prefer a shared causal kernel over superficial form differences. Item count, copy, imagery,
  page scale, optional flourish, or list-versus-single-item presentation may be adaptable variants
  inside one family when the controlling behavior and feedback loop are the same.
- Prefer one best child. Select two only when the complete Skill independently satisfies both
  routing boundaries; shared mood, subject, page type, or implementation is insufficient.
- Use `keep_at_parent=true` only when none of the discovered children can retrieve the Skill
  without distorting its meaning. Do not use the parent as an uncertainty fallback.
- Return every supplied Skill exactly once. Do not invent or rewrite names.

Return exactly JSON:
{"assignments":[{"skill_name":"exact supplied name","child_names":["exact supplied child"],"keep_at_parent":false,"reason":"brief mechanism-level reason"}]}
JSON only.
\end{promptbox}

\subsection{Augmenting User Instructions with Visual Skills for Plug-and-Play Website Generation}
For plug-and-play website generation, WebViSE augments an original user instruction with retrieved Visual Skills. Skill retrieval reuses the recursive routing Prompt~\ref{prompt:dag-routing} and the full-document applicability-review Prompt~\ref{prompt:skill-applicability-review}; the reviewed Skills are then passed to query expansion and consistency review below.

\subsubsection{Query expansion with reviewed Visual Skills}
\label{prompt:query-expansion}
\begin{itemize}[leftmargin=3mm]
    \item \textbf{Input:} The original website-development Query and the complete documents of the reviewed top-$K$ Visual Skills.
    \item \textbf{Output:} A JSON object containing only the compatible visual-effects requirements to append to the original Query.
\end{itemize}
\begin{promptbox}{Query expansion with reviewed Visual Skills}
Write the visual-effects requirements to append to one website-development Query using only the supplied, already-reviewed Visual Skill documents.

The original Query remains authoritative. Transfer only compatible observable composition, motion, interaction, state change, and feedback from the supplied Skills. Adapt mechanisms to the Query's existing subject and task; omit source branding, copy, assets, frameworks, and unrelated workflow. Combine complementary mechanisms coherently and avoid redundant demonstrations or one-section-per-Skill assembly. Do not mention Skills, retrieval, sources, or this review.

Return exactly one JSON object:
{"query":"visual-effects requirements section only"}

Use the original Query's natural language. Return only the JSON object.
\end{promptbox}

\subsubsection{Visual-effects consistency review}
\label{prompt:visual-effects-consistency-review}
\begin{itemize}[leftmargin=3mm]
    \item \textbf{Input:} The unchanged original website-development Query and the visual-effects requirements produced from reviewed Visual Skills.
    \item \textbf{Output:} A JSON consistency decision, a list of identified issues, and the unchanged or minimally repaired visual-effects section; the result may be empty only when no addition is compatible.
\end{itemize}
\begin{promptbox}{Visual-effects consistency review}
Review whether an appended visual-effects section is compatible with its original website-development Query.

The original Query is authoritative and remains unchanged outside this review. The appended section may add concrete visual composition, motion, interaction, state changes and feedback, but it must not restate, change or erase the original meaning.

An explicitly open creative brief is a deliberate exception to the usual no-new-subject rule only when it positively delegates the subject, concept and design direction to the creator (for example, “build whatever you want” or “express yourself without restrictions”). For that case, an invented coherent premise, content structure and many compatible spectacular effects are consistent with the original intent and should not be removed merely because the original supplied no subject. Check instead that the additions form one buildable creative work rather than unrelated demonstrations. Ordinary underspecified Queries do not receive this exception.

Treat an explicit source-fidelity constraint such as “1:1 replicate”, “exact clone”, “pixel-perfect reproduction”, or genuinely equivalent wording as authoritative. A bare request for a clone, counterpart or site in the style of a named product is not by itself an exact-fidelity constraint: it establishes the product domain, recognizable information architecture and core task pattern, while still allowing compatible authored visual mechanisms that serve those same subjects and tasks. Do not claim that an added mechanism exists in the named source, and do not change the referenced product into a different kind of experience. Reserve the strict no-invention rule for language that positively demands exact reproduction. Under that strict rule, remove unsupported additions even when they seem stylistically plausible; if no appended requirement can be retained without risking divergence from the explicitly exact source, return an empty final_query.

Check the subject or entity being presented, who owns the page, the speaker and addressee, pronoun and deictic references, intended audience, page goal, user tasks, required content and functionality, and output language. Interpret second-person references in the original as addressed to the model receiving the website-building request unless the original explicitly establishes another referent.

Do not audit or enforce programming languages, frameworks, styling systems, libraries, file formats or delivery modes in this visual-consistency pass. They belong to the unchanged original Query and must not be copied into, used to weaken, or added as qualifiers to the visual-effects section. Judge whether the visual intent is semantically compatible with the requested subject and task, not whether a particular stack can implement it. Keep `final_query` technology-agnostic and never prepend, quote or repeat the original Query.

Also check how the effects are composed. Distinct experience regions are reasonable when the Query is organised around presentation, exhibition, editorial progression or exploration. For a simulation, tool, generator or control-oriented Query, prefer effects that coherently support the core subject, shared state and task where sensible, while still allowing useful inspection, explanation, diagnostic or supporting regions. A visually sparse or compact layout does not imply few accepted effects: several Skills may be fused into one stage as complementary layers, for example subject material, motion, direct manipulation, impact feedback and atmosphere. Reject unnecessary vertical chaptering and unrelated demonstrations, not a well-integrated multi-Skill composition. Do not enforce one universal arrangement. Repair an effect when it replaces a concrete requested subject with an abstract visual object, obscures the subject's recognisable material or causal behaviour, or creates an unrelated demonstration instead of serving the Query.

This is also a completeness and usability audit. Check whether the section visibly covers an explicitly requested action or causal process; whether a dashboard, tool, editor or simulation exposes meaningful existing variables rather than freezing them; whether current/default time remains distinct from an explorable date or range; whether long overlays remain scrollable with reachable exit controls and restored focus/scroll; whether a concrete subject has been replaced by a generic metaphor; and whether an unsupported dark or low-luminance default has been introduced. These are repairable issues even when the added effect does not literally contradict a sentence in the original Query. Do not add arbitrary new product functions, but controls needed to inspect the Query's existing data, state or causal behavior are not arbitrary.

Audit colour by visual role rather than forcing one palette over the whole page. Avoiding an unsupported black, low-luminance or neon default does not require a light-neutral, achromatic, pastel or low-saturation result. Never use beige, cream, ivory, off-white, paper, earthy, warm-neutral or another named "safe" palette merely as the replacement for an unsupported dark treatment. If the original Query neither specifies a palette nor makes one necessary to the subject, do not prescribe a fixed palette at all: state only the required colour roles, contrast, legibility and distinction, and leave the concrete art direction unspecified. Preserve a named palette only when the original explicitly requests it or the subject intrinsically requires faithful colours. Treat any unsupported named palette introduced by the appended section as an inconsistency and remove or repair it, including unsupported light and warm-neutral palettes, not only dark ones. When patterns, generated artifacts, illustrations, media, data identities, materials or other visual subjects are central to the original Query, a blanket neutral-palette instruction that flattens their variety or expressive role is an unsupported restriction and a visual-completeness failure. Keep structural ground, typography and UI chrome coherent and accessible, while allowing or requiring the subject-defining visuals to use purposeful colour diversity, contrast and saturation when the original Query does not forbid it. Conversely, do not inject arbitrary colour variety when the original explicitly fixes a restrained palette. Repair an over-broad palette clause at its scope rather than replacing it with another house style.

Also audit the visual ambition of an underspecified creative surface. When a generator, creator or customizer produces an artifact whose appearance is central and the original Query does not lock one style, a single fixed treatment plus file-format choices is incomplete: require several materially different visual presets or meaningful appearance parameters, a live preview, a persistent selected state, and an exported artifact that matches the preview while preserving the artifact's validity. These are visual treatments of the requested output, not new business functions. For simulations and animation-led experiences whose art direction is not fixed, retain or minimally repair a coherent distinctive visual language through compatible material response, trajectory or trail, collision/reaction feedback, spatial depth or atmosphere; controls alone do not satisfy this audit. Do not require every possible effect and do not add a generic dark neon theme. Prefer one recognisable, coherent visual signature over a pile of unrelated decoration.

If the section is both consistent and complete, preserve it verbatim. Otherwise repair only the conflicting, missing-core, frozen-control, accessibility or usability requirement. Do not rewrite or repeat the original Query. Retain useful compatible visual mechanisms, and prefer deleting or narrowing a conflicting clause over discarding an otherwise compatible section. Lack of evidence that a compatible authored mechanism appears in a named reference is not, by itself, a reason to erase it unless the Query explicitly requires exact fidelity. When the section still contains any visual effects after repair, keep at least one suitable effect on the home view or first visible screen. Do not force an effect to survive: final_query may be the empty string only when every addition conflicts with the original intent or an explicit exact-fidelity constraint.

Use exactly:
{
  "intent_consistent": true,
  "issues": [],
  "final_query": "the unchanged or minimally repaired expanded Query, or an empty string when no addition is safe"
}

Return only the JSON object.
\end{promptbox}

\end{document}